# Challenges in annotations by humans and LLMs: A case study of evaluative language

**Mirela Imamovic**[1]
University of Hildesheim, Institute of Translation Studies and Specialised Communication, Germany
University of Paris 8 Vincennes-Saint Denis, TransCrit, France
imamovicm@uni-hildesheim.de
https://orcid.org/0000-0002-9064-5771

**Aenne Cecilia Kristine Knierim**
University of Hildesheim, Institute of Information Science and Language Technology, Germany
knierim@uni-hildesheim.de
https://orcid.org/0009-0006-1584-0717

**Khushi Pitroda**
University of Hildesheim, Institute of Translation Studies and Specialised Communication, Germany
pitroda@uni-hildesheim.de
https://orcid.org/0009-0008-0629-5432

**Ekaterina Lapshinova-Koltunski**
University of Hildesheim, Institute of Translation Studies and Specialised Communication, Germany
lapshinovakoltun@uni-hildesheim.de
https://orcid.org/0000-0002-5618-8087

**Abstract**
In this paper, we draw a comparison between linguists in training, a trained linguist, and annotations generated by large language models (LLMs) to find out if they struggle with complex linguistic phenomena in a similar way. For this purpose, we analyse evaluative language in spoken popular science discourse, with the example of a corpus of English TED talk transcripts. We focus on the Appraisal theory and its Attitude subsystem, including the categories (classes) of Affect, Judgement, and Appreciation. In this context, Appraisal theory is an example of a highly subjective annotation task, making it a suitable example for the study of complex annotation challenges. First, we assess human annotations on a sentence level in specific scientific domains. Then, we develop three prompts and compare them for model performance for the automatic classification of Appraisal classes. We assess the performance of three LLMs using the best-performing prompt and finetune the model, reaching an F1-score of 0.77. We find that models perform best compared to annotations conducted by the trained linguist, while linguists in training do not reach high agreement scores. We conclude that LLMs can aid in complex annotation task resolution, opening new pathways for the complex theories annotated and analyzed in digital humanities studies.

**Keywords:** Evaluative language, Appraisal theory, Large language models, Complex annotation task resolution, Machine-supported annotation, TED talks

## 1 Introduction

The rapid evolution of natural language processing (NLP) has substantially advanced text-centered research in the digital humanities (DH) as well as in computational social sciences (CSS). Until recently, supervised learning approaches have dominated the field, requiring large amounts of labelled data to train language models. Although large language models (LLMs) have demonstrated strong performance on many NLP tasks in an unsupervised manner, they do not consistently outperform supervised machine learning models. Consequently, the need for high-quality training data has not become obsolete. For example, in stance detection – a task closely related to Appraisal identification – performance accuracy varies considerably across models and prompt designs (Cruickshank et al., 2025). Moreover, supervised models offer advantages such as domain specificity.

In addition, the complexity of the theories and concepts being annotated is increasing. Tasks such as identifying a cat in an image or determining whether *swim* is a verb typically yield high agreement scores, whereas detecting toxicity in a text or meme is far more subjective and complex

[1] Both Mirela Imamovic and Aenne C. K. Knierim contributed equally and share first authorship.

(Plank, 2022). In contemporary interdisciplinary fields such as DH and CSS, increasingly nuanced theoretical constructs are being operationalized and tested, including concepts such as solidarity (Kneuer et al., 2025). In many cases, a trade-off has to be made between the complexity of the concept being annotated and the feasibility of the classification tasks (Knierim & Heid, 2025). Given the strong performance of LLMs across many NLP tasks, researchers are increasingly exploring the opportunities afforded by LLMs for a broad range of applications. Concerning annotation practices, applying LLMs to complex annotation processes may enhance the availability of structured data (Ziems et al., 2024). For example, codifying twenty-two rhetorical and linguistic features to identify propaganda techniques, Hamilton et al. (2024) demonstrate how combining a small number of human-annotated examples with GPT can be an effective strategy for 'scaling the annotation process at a fraction of the cost of traditional annotations relying solely on human experts'. However, it is necessary to validate the performance of LLMs for text annotation on a task-by-task basis, as the performance is highly contingent on the type of annotation task (Pangakis et al., 2023). In this paper, in line with Ziems et al. (2024), we aim to find out if LLMs match or exceed the performance of human annotators, even in complex tasks or if they face the same problems as humans. For this, we conduct a case study on evaluative language.

Evaluative language, or the linguistic expression of the speaker/writer's opinion and feelings (e.g. *beautiful painting, huge despair*), is a fundamental function of language and it has been extensively studied in the literature under different umbrella terms such as stance (Biber & Finegan, 1988; Biber & Finegan, 1989; Conrad & Biber, 2000), modality (Lyons, 1997; Palmer, 2014), evidentiality (Aikhenvald, 2004; Diewald & Smirnova, 2010), subjectivity (Langacker, 1985; Kärkkäinen, 2006; Englebretson, 2007), evaluation (Hunston & Thompson, 2000; Bednarek, 2006; Hunston, 2010; Thompson & Alba-Juez, 2014; Glynn & Sjölin, 2015) and Appraisal (Martin & White, 2005). We focus on Appraisal theory (Martin & White, 2005), a functional model and a detailed description of evaluative language, which was developed under the framework of Systemic Functional Linguistics (Martin, 1992; Halliday, 2004/1994; Halliday & Matthiessen, 2014). It consists of three interrelated subsystems: Attitude (descriptions of feelings and attitudinal assessments), Graduation (amplification or downtoning of evaluative meaning) and Engagement (interaction with other voices and perspectives in the discourse). These three subsystems further consist of several other categories and subcategories. We limit our analysis to the Attitude subsystem and its semantic/pragmatic categories of Affect, Judgement, and Appreciation.

Appraisal theory has been applied extensively in discourse studies (Eggins & Slade, 1997; Dong & Fang, 2023;), translation studies (Munday, 2012; Alsina et al., 2017; Imamovic et al., 2025), corpus linguistics (Bednarek, 2008; Cavasso & Taboada, 2021) and computational linguistics (Taboada & Grieve, 2004; Pang & Lee, 2008; Bloom, 2011). One advantage of Appraisal theory is that it is a valuable and widely used framework for capturing and describing a multifaceted phenomenon of evaluation in language. However, due to its richness, complexity and hierarchical nature, it poses several problems, especially when it comes to achieving high IAA scores and applying it in automatic approaches to extract Appraisal items. Some of its disadvantages are the complexity and difficulty of operationalization. Even though Martin & White (2005, pp. 58-59) propose the following linguistic frames that can be used to detect the expressions of Attitude: Affect (person feels affect about something), Judgement (it was judgement for person/of person to do that), Appreciation (person considers something appreciation), they remain inadequate for the precise identification and classification of Appraisal both in manual and automatic approaches.

Annotation remains a crucial process for capturing evaluative meaning in text. The identification and classification of complex pragmatic phenomena in text are two important steps for any linguistic research. However, several challenges have been mentioned in the existing studies that make this process both tedious and complex. For example, problems that many researchers working with Appraisal face are (a) identifying Appraisal-bearing expressions, i.e. deciding on the span of evaluation, and (b) distinguishing between subtle categories, especially when annotating different fine-grained ones. Manual annotation is the traditional way of identifying evaluative items in text; however, this process is very time-consuming and prone to subjective interpretation (Fuoli, 2018). Also, evaluative language is context-dependent, and the categories of Judgement and Appreciation are often mistaken because evaluative meaning usually associated with one lexical item can realise the other (Bednarek, 2009; Taboada & Carretero, 2012; Fuoli, 2018).

To add to this question, we conduct a case study, analyzing TED talks, a genre of spoken popular science and specialized communication, where experts speak and transfer information to non-experts in the field. We explore whether LLMs encounter similar challenges in classifying evaluative expressions as humans do. For this purpose, a corpus of English TED talk transcripts from eleven domains is used, containing expressions of positive and negative, explicit and implicit evaluative items. This study addresses the following research questions:

RQ1: Which Attitude categories are challenging for human annotators and what are the underlying problems?

RQ2: Using the example of Appraisal theory, do LLMs meet the expectations expressed in the scientific community related to complex annotation task resolution?

RQ3: Do humans and LLMs struggle with the same issues in the annotation of categories under analysis?

Our research contributions are twofold: First, we present an annotation scheme for Appraisal theory and its Attitude categories. Second, we contrast the performances of humans and LLMs in complex annotation tasks, exploring pathways for interdisciplinary research. We aim to demonstrate whether future studies can be operationalized within this realm and whether extensive annotated corpora are necessary for automatically classifying large portions of speech data, especially considering the Appraisal categories. If LLMs demonstrate strong performance in annotating complex linguistic theories through prompting, this could open new avenues for research in linguistics and the humanities. The remainder of the paper is organised as follows. We present the theoretical framework and related studies in Section 2. The details on our methodology are given in Section 3. While RQ1 is addressed in Section 4, we follow with RQ2 and RQ3 in Sections 5 and 6, respectively. Section 7 concludes, and Section 8 contains a discussion of problematic issues and limitations, as well as the following steps.

## 2 Appraisal Theory for the Annotation of Evaluative Language

Previous studies annotated evaluative language using Appraisal theory in popular science discourse and TED talks by experimenting with LLMs. For example, prompting strategies were optimised for the specific task of Appraisal annotation by using GPT-3.5, GPT-4o and Gemini 1.0 Pro (Imamovic, 2026). The few-shot prompting strategy was the most efficient, with GPT-3.5 achieving a Kappa score of 0.65 in comparison with human annotators. Also, ChatGPT was used to annotate Attitude categories, adopting a single-token approach (Imamovic et al., 2024). Like this, multiple evaluative words per sentence can be identified and classified. Their findings indicate that ChatGPT had a high precision in the identification of evaluative items, but also failed to identify several of them due to low recall.

Appraisal theory has also been applied to the analysis of other complex pragmatic phenomena, such as satire, irony, apologies and metaphor. For instance, Stingo & Delmonte (2016) annotated satire in Italian political commentaries made by prominent Italian journalists. The authors developed lexical, semantic and syntactic criteria for Appraisal annotation and allowed for the labelling of entire clauses and sentences. While they provided extensive information on the annotation criteria, they did not report IAA scores. Similarly, Busetto & Delmonte (2019) analysed irony in 154 of Shakespeare's sonnets by manually annotating the Attitude categories and their polarity in the corpus. Although they did not report detailed agreement, they mentioned that the first author annotated the corpus, while the second author reviewed it for discrepancies, resulting in approximately 65% agreement. In another study, Yu et al. (2024) compared LLMs (GPT-3.5 and GPT-4) with a human annotator in the annotation of apologies in English. They reported promising results from the automatic pragmatic annotation process. Another study (Fuoli et al., 2022) contributed new annotation guidelines for creative metaphors using Appraisal theory in film reviews. They reported a mean PABAK agreement score of 0.69. However, among the labels used, the involved (implicit) evaluation showed the lowest agreement. The authors attributed this to the inherently subjective and context-dependent nature of such evaluations.

Previous studies have also analysed the other two subsystems of Appraisal, Engagement and Graduation. For example, Dong and Fang (2023) annotated Engagement in two comparable English media corpora. They used different media types and focused on text categories of news, editorial, society, culture and arts, and business. While the authors proposed a procedure to improve annotation, they did not report any agreement scores. In another study, Taboada et al. (2014) took a contrastive approach and analysed Attitude and Graduation in the corpus of movie reviews in English, German, and Spanish. Similarly, Read & Carroll (2012) analysed a corpus of book reviews, experimenting with different annotation strategies, such as single token, multi-word expression and sentence-level annotation. In the end, they allowed for a free annotation of multiple Appraisal tokens per sentence. Two annotators worked in two rounds and discussed the cases of disagreement halfway through. The annotators achieved a mean F-score of 0.698 in identifying Appraisal expressions (span agreement). For the classification between different Appraisal subsystems (Attitude, Engagement and Graduation), a mean F-score of 0.635 was reached. Among the Attitude categories, the annotators achieved a mean F-score of 0.519 for Affect, 0.586 for Judgement and 0.567 for Appreciation. The authors reported on the shortcomings of their annotations scheme in that it allowed single Attitude category per linguistic item, stating that: ‘words may be relevant to multiple classes; the class of a word can vary according to its context; appraisals may relate to multiple entities [...]; and interpretation of Appraisal-bearing words is subjective’ (Read & Carroll 2012, p. 444). Their annotators disagreed on several instances, offering plausible reasons why another category would fit better.

Other authors also acknowledged the challenges involved in annotating Attitude categories within corpora of news media editorials (Zeng et al., 2024). These challenges often stem from ambiguities between categories, particularly between Affect and Judgement, as well as Affect and Appreciation. Aiming to reduce inconsistencies, they proposed using a single-label annotation approach rather than a multi-label one. In our study, we allow multi-labelling but include the Ambiguous category to investigate whether this helps to show which categories are difficult to distinguish when they simultaneously express multiple evaluative meanings. Toprak et al. (2010) analysed consumer reviews annotated on opinion-related information using a sentence-level (an opinion of a review topic) and expression-level (target of evaluation, modifiers) approach. Even though they do not use an Appraisal scheme, they annotate explicit expressions of opinions in terms of an author expressing an attitude towards a target of evaluation, as in *I had a nightmare with Capella University* (p. 3) and implicit expressions of opinion that are stated as facts but do offer an evaluation of the entity mentioned. In this study, we also use a sentence-level approach in the classification of Appraisal categories.

## 3 Corpus Compilation

TED talks, a genre of scientific presentations, exemplify popular science discourse as they are delivered by expert speakers and aimed at a general audience. Since they represent a form of specialized communication in which experts address non-experts, TED talks are particularly well-suited for analyzing evaluative language due to their inherently persuasive nature. Therefore, we can expect to find evaluative expressions throughout their talks, as they aim to engage the audience emotionally, build credibility, and persuade listeners to adopt a certain viewpoint or take specific steps. These expressions can help us understand how the speakers position themselves, their topics, and relate to their audience within the discourse.

The data were manually collected from the TED website[2] in March and April 2023. The corpus, called Emotionaliz*TED*[3], contains 190 English texts. The eleven domains within which speakers delivered their talks are shown in Table 1. We select 22 English transcripts (two per domain) from the bigger corpus and let them be manually classified by linguists in training. Each transcript set is balanced in terms of the speakers’ gender; for each domain, we include two talks, one by a male speaker and one by a female speaker. Table 1 provides an overview of the number of texts per domain, as well as descriptive information such as document size and type-token ratio.

| Domain | Number of texts (N=190) | Token count | ⌀ Sentences per text | TTR |
|---|---|---|---|---|
| Art | 16 | 35859 | ~ 141 | 0.131 |
| Business | 16 | 32695 | ~ 134 | 0.132 |
| Education | 15 | 37084 | ~ 141 | 0.133 |
| Entertainment | 15 | 30211 | ~ 135 | 0.128 |
| History | 15 | 29321 | ~ 114 | 0.157 |
| Medicine | 17 | 41717 | ~ 147 | 0.119 |
| Science | 22 | 55576 | ~ 152 | 0.099 |
| Philosophy | 15 | 41709 | ~ 177 | 0.122 |
| Politics | 17 | 42622 | ~ 146 | 0.122 |
| Psychology | 17 | 39553 | ~ 148 | 0.114 |
| Technology | 25 | 43221 | ~ 101 | 0.122 |

**Table 1** Simple corpus statistics for the Emotionaliz*TED* corpus, including type-token-ratio (TTR)

## 4 Manual Annotation: Process, Guidelines, and Analysis

[2] https://www.ted.com/

[3] The corpus of texts used for this study can be accessed on this GitHub page: github.com/anonymous/

In this section, we introduce the guidelines for the manual annotation and report on the annotation process of human annotators. We present results of manual annotation using Cohen's Kappa (Cohen, 1960) and Krippendorff's alpha (Krippendorff, 2011). Next, we conduct an error analysis to answer our first research question: Which Attitude categories are challenging for human annotators and what are the underlying problems?

### 4.1 Annotation Process and Guidelines

The annotations for this study were done in a class throughout the course of one semester. The students, who are linguists and translators in training, received training in Appraisal theory, familiarizing themselves with the Attitude and Graduation subsystems of Appraisal[4]. They regularly completed homework and additionally practised annotation in pairs in the class. Due to the strict class setting and lack of time, annotation iterations (as suggested by Fuoli, 2018) in terms of discussions between annotators were not conducted. However, difficult examples were discussed in class together. The time spent on the annotation practice in class was considered sufficient for the annotators to be equipped with enough knowledge and perform the classification of evaluative meanings. Since training and recruiting annotators can be a costly endeavour, opting for annotations by students in training is a practical solution.

For our annotation task, we employ twenty-four student annotators enrolled in a BA programme in Linguistics at Anonymised. The majority of the annotators are native German speakers (n = 15), alongside some Spanish Erasmus students (n = 6), two Russian and one Turkish Erasmus student. While the annotators also have academic training in translation studies, they are not native speakers of English. This might have influenced the Inter-Annotator agreement, as the annotation of Appraisal requires nuanced and specialized linguistic competence typically associated with native-level proficiency (Read and Carroll, 2012). However, several students do have C1/C2 English level skills. Even though we do not include controls in our study design (e.g. comparison with a group of annotators who did not receive training in Appraisal theory), we employ a senior researcher and a linguist with experience in working with Appraisal theory and also a non-native English speaker with a C2 English level. The senior researcher annotated a random sample of sentences from the corpus (twenty sentence examples from each text, amounting to forty sentences per domain and 440 sentence examples in total). We consider this as the gold standard for the task.

Each linguist in training was assigned two TED talks from a specific domain, which made it possible to have annotations by two annotators for each text. The annotations were done in two rounds to ensure the guidelines were understood correctly. The annotators were instructed to classify each sentence of a TED talk, considering the overall evaluative meaning expressed in the sentence and choose corresponding Attitude categories. Based on the TED speaker's opinion and assessments expressed in the talk, they had to interpret how the sentence as a whole evaluates people, things, or expresses the speaker's feelings. We decided to operate on the sentence level in order to avoid the problem of the item scope identification, i.e. determining the borders of the evaluative expression (Imamovic et al., 2024), since we noticed that the annotators struggled with this. A sentence-level approach is also used in existing works on sentiment analysis, but since it has never been used in studies on Appraisal (as far as it is known to us), we decided to experiment with it.

When it comes to the annotation guidelines, the annotators had to classify if the sentence was evaluative or not, i.e. if the sentence expresses an emotion, judgement (ethics) or aesthetic evaluation. If the sentence was purely factual or neutral, annotators were asked to classify it as non-evaluative. We briefly introduce the classes (Attitude categories) to be annotated and how they were operationalized. The detailed annotation guidelines are displayed in Appendix 3. For the annotation of Attitude, we used the following classes: Affect (emotions), Judgement (ethics), and Appreciation (aesthetics). Affect was supposed to be used if the sentence expresses feelings or emotions (e.g. *happy, scared, grateful*), Judgement if it evaluates behaviour or character of people (e.g. *kind, unfair, brave*) and Appreciation if things, performance, or phenomena (e.g. *beautiful, boring, excellent*) were evaluated. We also included two more classes, Ambiguous and Uncertain. Ambiguous was supposed to be used if more than one Attitude category applies, but it is unclear which. The annotators also had to specify the categories they were considering as ambiguous (e.g. Affect, Judgement). They were instructed to use the label Uncertain only if necessary. We explained that it leads to a loss of usable data for model training and that they should only use it if not confident about their choice. The annotators were allowed to choose more than one Attitude category (e.g. Affect and Judgment), i.e. multi-label annotation was allowed, provided that they were confident that different evaluative meanings were expressed. This is because one sentence can express Appraisal both explicitly and implicitly, as in e.g. *Despite everything, she stood tall and smiled.* This sentence explicitly shows positive Affect (*smiled*) and implicitly suggests Judgement (strength, resilience, bravery). The annotators were asked to pay attention to the co-text (sentences before and after the one they were classifying) to grasp the full meaning. Fig. 1 below presents the Appraisal resources as proposed by Martin & White (2005).

[4] The Engagement subsystem was left out because of its hierarchical complexity.

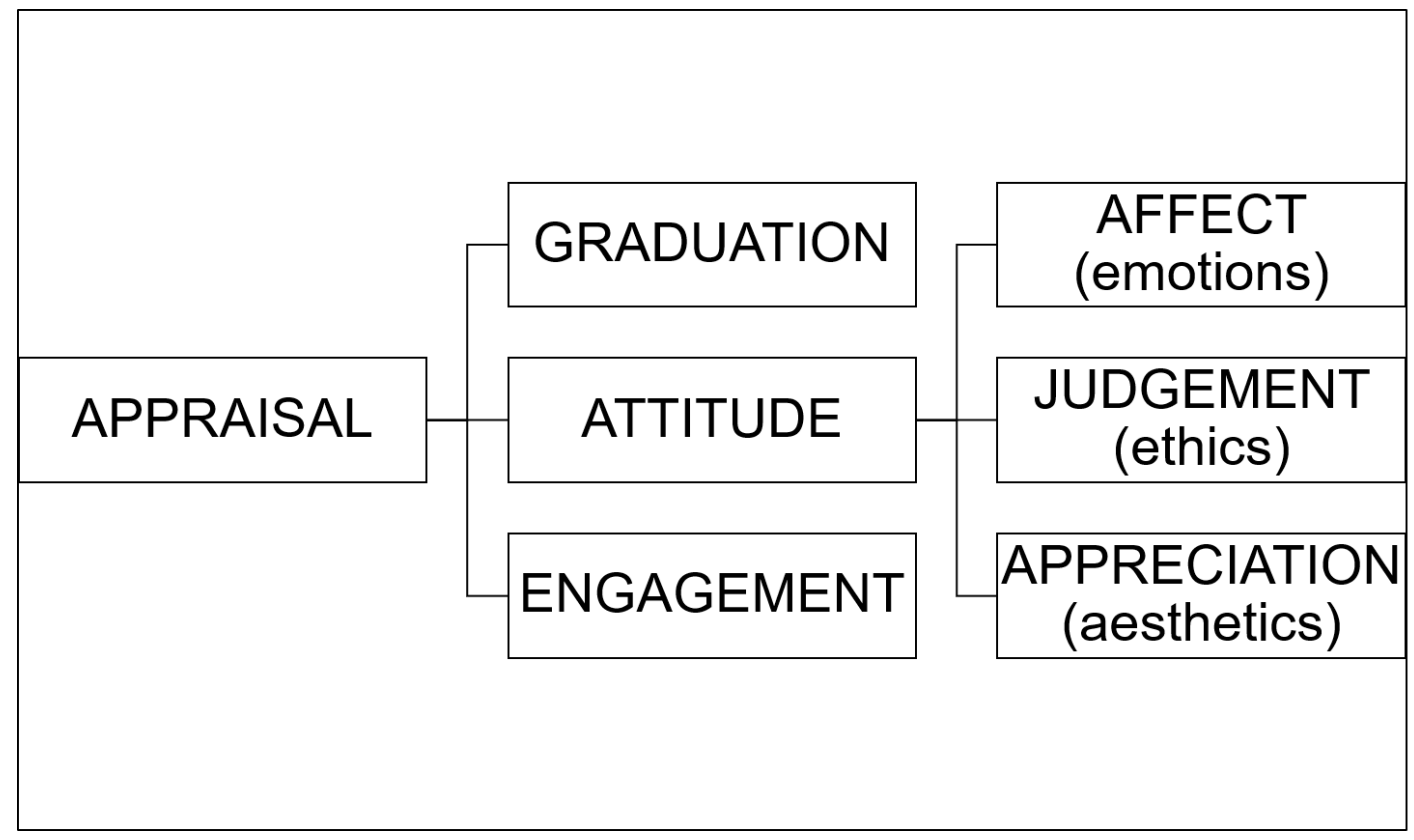

**Fig. 1** Appraisal resources based on Martin & White (2005). The figure shows a tree-structured graph with Appraisal as the parent node. The Appraisal subsystems Graduation, Attitude, and Engagement are shown. The Attitude system is diversified further into the subcategories Affect (emotions), Judgement (ethics) and Appreciation (aesthetics)

### 4.2 Human Annotation Evaluation

We first proceed with the evaluation of the annotations performed by the linguists in training. Before choosing one of the possible labels, Affect, Judgement, or Appreciation, annotators were to decide whether a sentence is evaluative or not. A sentence is considered to be evaluative if both annotators agree that it is evaluative. To calculate the IAA for the agreement on sentence evaluativeness between the linguists in training only, we use Cohen's Kappa. For all other calculations of agreement scores, we use Krippendorff's alpha, as it allows for multiple annotators and more than two classes. Interestingly, IAA varies across texts across different domains. For some domains, i.e. Entertainment, Politics, Education, Technology, the agreements are moderate (above 0.50). However, agreement drops sharply for History and Psychology, followed by Medicine and Business texts (see Fig. 2).

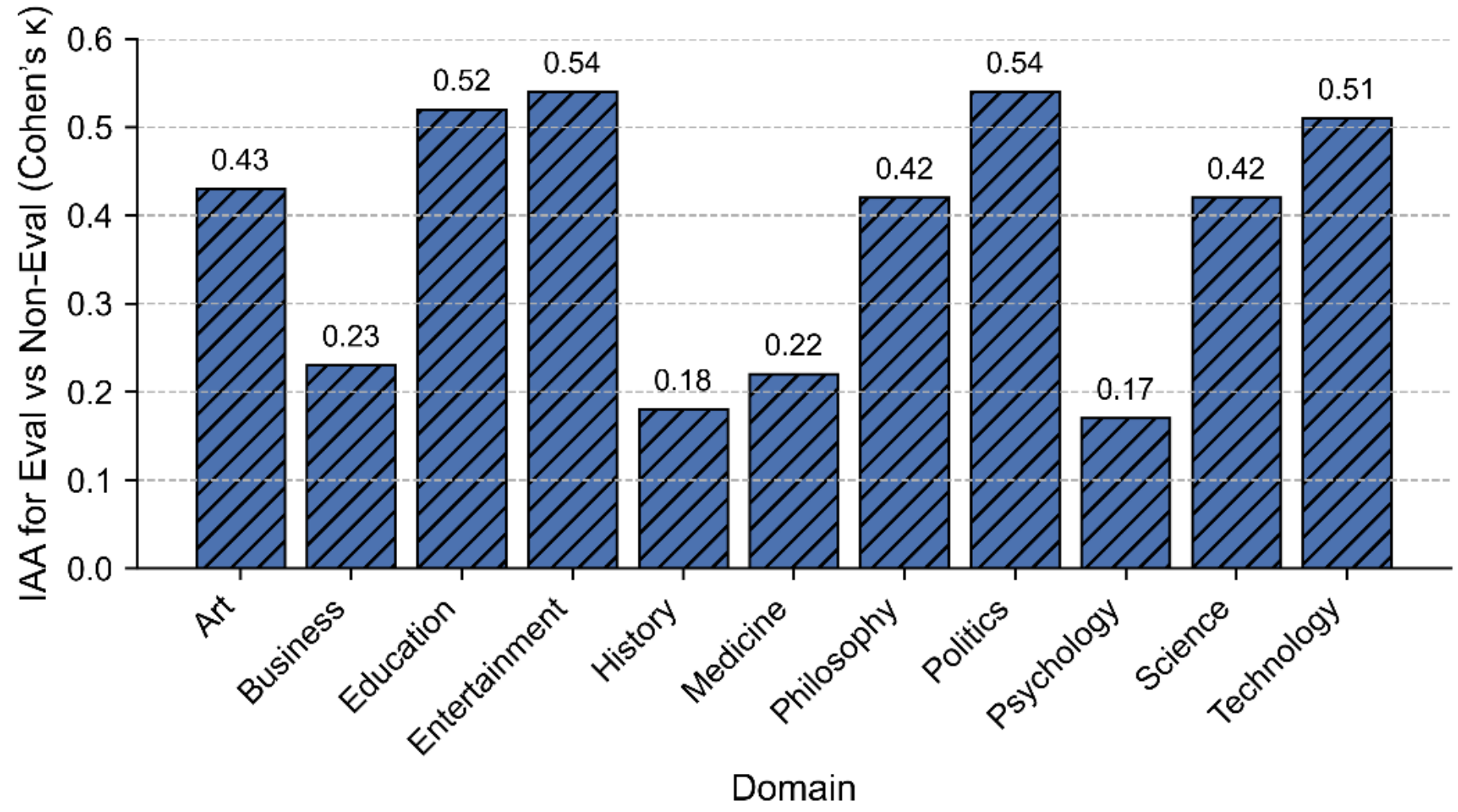

**Fig. 2** Cohen's Kappa for the IAA between the linguists in training for the evaluativeness of the sentence. The bar chart shows the IAA per domain. For the domains Education, Entertainment, Politics and Technology, annotators reach the highest agreement with κ >= 0.51. For the domains Art, Philosophy and Science, moderate agreement

is reached. Annotators agree fairly for the domains Business and Medicine and agree slightly for the domains History and Psychology

After observing IAA for the evaluative class, we next look into the agreement scores for the classification of the three subclasses (Attitude categories) for the evaluative sentences. Expectedly, agreement for the three subclasses is much lower (see Fig. 3) than the agreement in the identification of evaluative sentences (see Fig. 2). Variation across domains is observed here, too. The worst results are achieved for History and Psychology. However, it is interesting to see that a good agreement for specific classes seems to be dependent on the discipline under analysis (we elaborate more on this in Section 4.4). Thus, the highest agreement in Affect is achieved in Business, Education, Entertainment, Philosophy, Politics, and Science, whereas the highest agreement in Appreciation is observed for Entertainment and Arts. Agreement for the category of Judgement is better in Education, Medicine, and Technology.

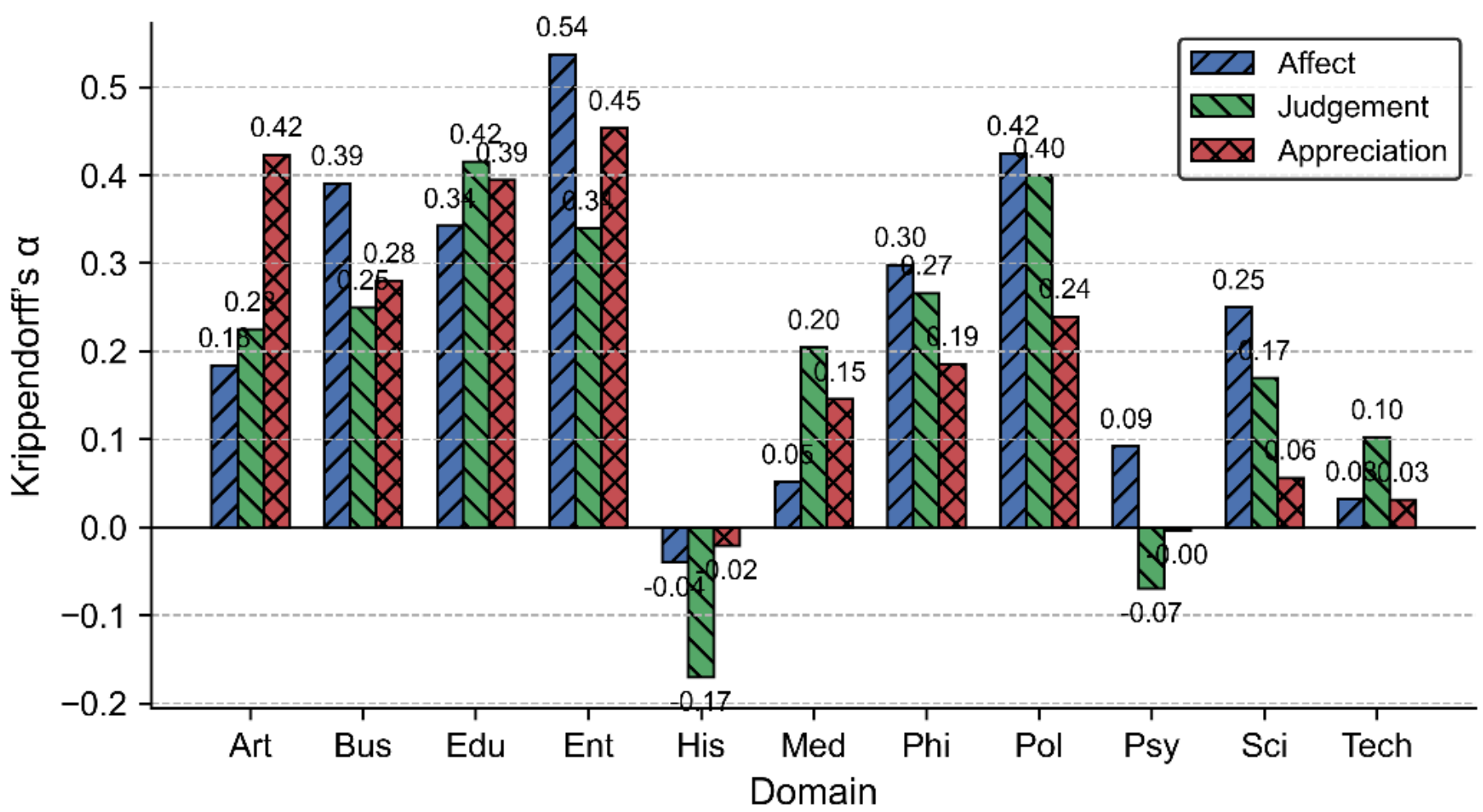

**Fig. 3** This graph shows the IAA between the linguists in training for the labelling of the subclasses Affect, Judgement, and Appreciation. It is a stacked bar chart with three bars per domain, representing the three subclasses. The Inter-Annotator agreement is calculated using Krippendorff's alpha. Agreement for Appreciation ranges between α = -0.02 and α = 0.45. The agreement for Judgement lies between α = -0.04 and α = 0.42. For Appreciation, annotators reach agreements between α = -0.02 and α = 0.54. Notably, the highest agreements on average are obtained for the domains Education, Entertainment and Politics

Next, we look at the agreement scores between the linguists in training and a trained linguist. We also calculate the IAA for the agreement on sentence evaluativeness and for the three Attitude subclasses. Krippendorff's alpha is used here instead of Cohen's Kappa because we measure agreement between three annotators, not two. When we compare the agreement on the evaluativeness of the sentence, we see that the highest scores are obtained for Technology, Education, Philosophy, and Science (see Fig. 4). The lowest score is obtained for Psychology, which is not surprising, considering that it was also low among the linguists in training themselves. However, interestingly, for the domain of Entertainment, the scores are also amongst the lowest, compared to the highest score achieved for this domain between linguists in training. This suggests that, even though the linguists in training made similar decisions on what is evaluative or not in that domain, this decision could be considered inaccurate if judged against the gold standard.

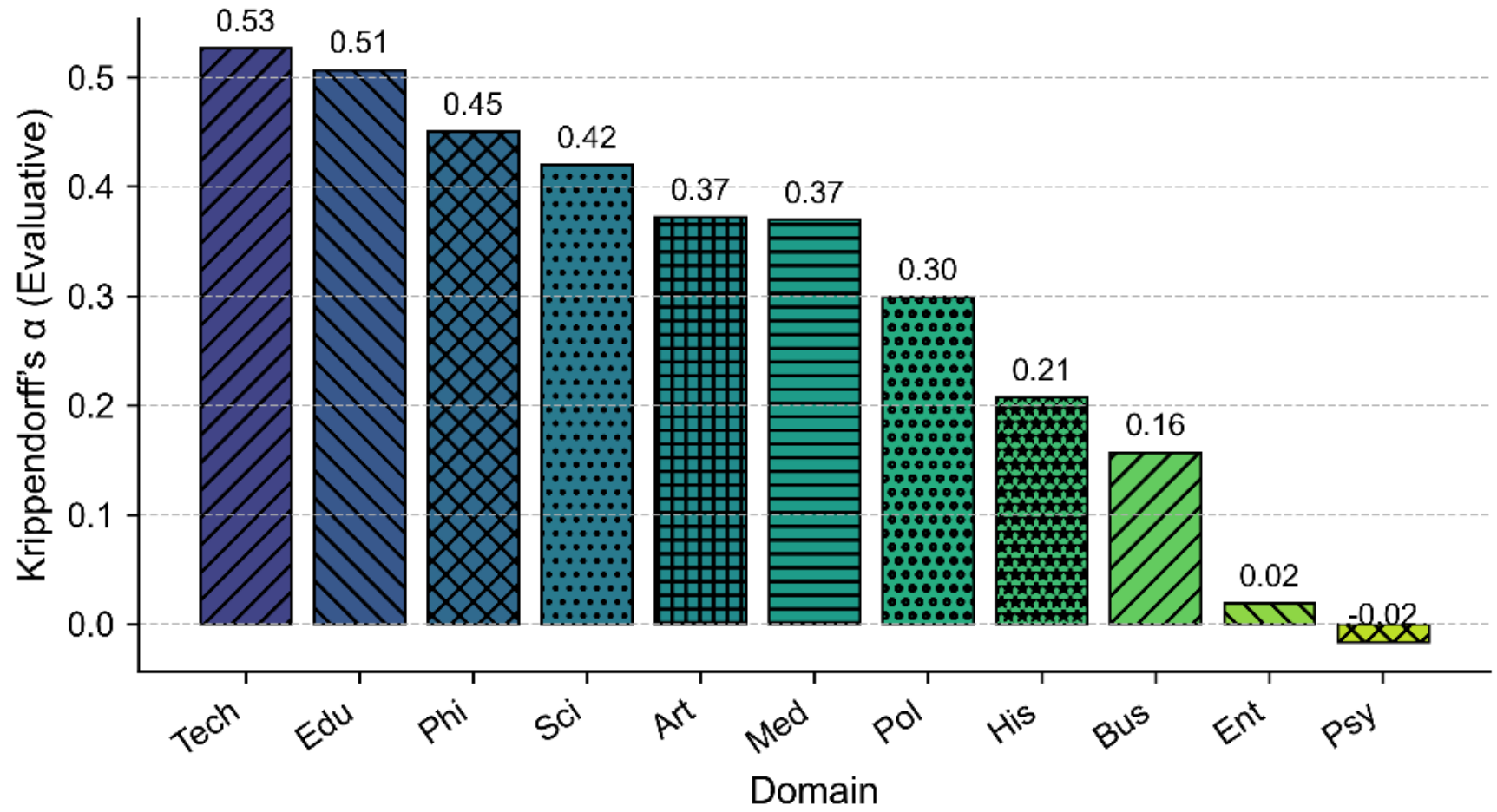

**Fig. 4** IAA for the evaluative class between senior researcher and linguists in training per domain. Moderate agreement is obtained for the domains Technology, Education, Philosophy and Science. For the domains Art, Medicine, Politics and History, annotators reach fair agreement. While there is no agreement for the domain of Psychology, slight agreement is reached for the domains of Business and Entertainment

We observe similar results for the domains of Politics (high agreement between linguists in training vs. agreement between them and the trained linguist) and Medicine (low agreement between linguists in training vs. agreement between them and the trained linguist), too. Comparing the agreement for the three subclasses (see Fig. 5), the highest one in Affect is achieved for Psychology and Business, followed by Education and Arts. We observe the highest agreement in Appreciation for Psychology, Entertainment, and Education. For Judgement, the highest scores are observed in Arts, Education, and Philosophy, and the worst in Business, Technology, and Medicine.

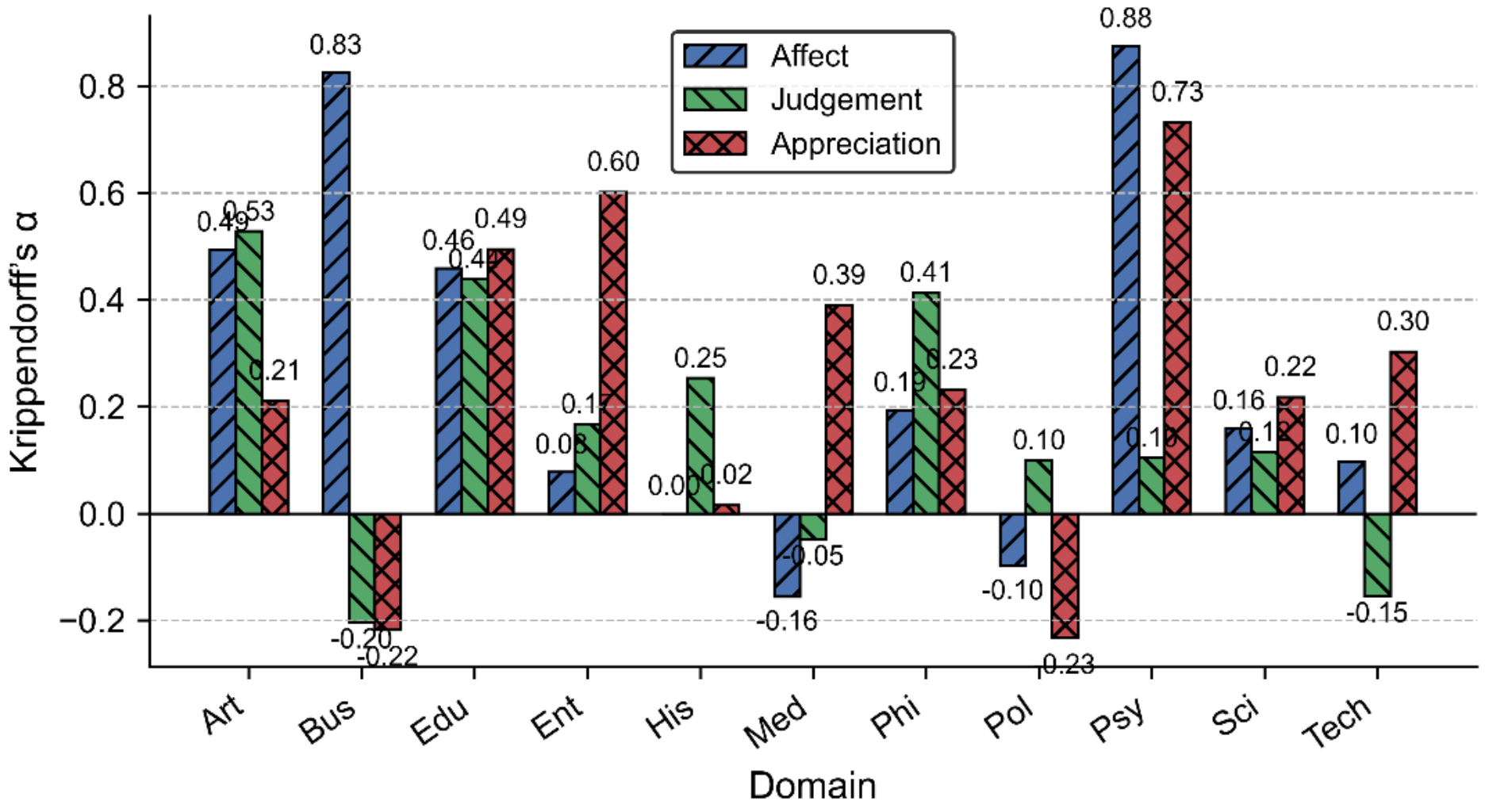

**Fig. 5** IAA between the senior researcher and linguists in training for the labelling of the subclasses Affect, Judgement, and Appreciation. This bar chart shows that higher agreement scores are obtained for Affect and Appreciation. For Business and Psychology, the annotators reach α >= 0.83, which is near perfect agreement. However, for many domains, agreement is only slight or fair. On average, the lowest agreement is reached for the Judgement class

### 4.3 Error Analysis: Challenges in Annotation faced by Annotators

In this section, we report on the results of the qualitative analysis. Given the high complexity and subjectivity of the task, we conducted an error analysis. We show agreement both among the linguists in training and the one between them and a senior researcher. The low IAA scores indicate that annotators struggle with annotating Appraisal. As mentioned in the previous section, at first glance, it seems that a high agreement between annotators depends on the domain they were assigned to work with. However, in this section, we show that the IAA scores depend on which annotators were considered for a specific text and domain, so the choice of students also affected the results and the agreement on the domains from which they were assigned a text. The low agreement in some domains (e.g. those under 0.50) is due to students' English language level and other factors such as interest in annotation and fatigue. However, another plausible reason could be that some students struggled with understanding the domain topic since TED talks are a specialized genre. Even though the type-token ratio of the texts is low and the language should have been understandable, specialized terminology and concepts might have been ungraspable. Ultimately, identifying evaluative expressions in TED talks might not be an easy task, as the talks are designed to be persuasive and contain many personal opinions and anecdotes.

To report on the challenges that the annotators faced, we proceed with the qualitative analysis of the annotations and focus on disagreements. First, we start with the binary classification, where the annotators had to decide whether the sentence expresses evaluation. Notably, the findings indicate that at least one annotator demonstrates difficulty in identifying whether a sentence is evaluative, even in cases where an explicit evaluative expression is present. This becomes apparent in the following example (1).

(1) *And I remember we got really **badly** lost.*

One annotator classified the sentence as non-evaluative, while the other classified it as evaluative with a subclass Ambiguous: Affect, Appreciation. The sentence in (1) is evaluative as it contains the adverb *badly*, which in this context explicitly indicates the speaker’s assessment of the state of being lost. While the sentence is evaluative, errors of this kind may occur due to fatigue or lapses in attention on the part of the annotators. If they were to work together in pairs and have post-annotation discussions (i.e. annotation iterations) after the classification of evaluativeness, the student who deemed the sentence as non-evaluative would quickly realise that they made a mistake, as the other student would point out that the sentence is evaluative. Another reason for this lapse could be that the associations with the used words in English are different depending on the native language of the annotators. In this specific example, the annotators had different native languages (German and Russian).

Another challenge in the classification of the evaluativeness of the sentence is the subjective interpretation of implicit meanings expressed in the sentence, where either the senior researcher or the annotator chooses an evaluative label and the other does not, resulting in different classifications. This occurs due to the subjective interpretation of the meaning expressed in the sentence. As shown in (2) below, the sentence was annotated as evaluative (Appreciation) by the senior researcher and annotator 1. However, annotator 2 classified it as non-evaluative. Even though the sentence looks purely factual, due to the subjective interpretation on the part of the given annotator, it might be deemed as implicitly offering an Appreciation, as evidenced by at least two annotators agreeing on it. Both linguists in training have an excellent command of English, which means that the misclassification for evaluativeness is not the misunderstanding of Appraisal on their behalf, but of failing to grasp the full meaning expressed (e.g. low concentration or fatigue). Another reason is not having discussions to express their opinions and find one solution together. Therefore, having post-annotation discussions is a plausible solution for subjective interpretations of implicit meanings, as annotating implicit meanings tends to contribute to low agreements, as shown by Fuoli et al. (2022).

(2) *We have knowledge.*

For the disagreement in the classification of Attitude categories, we identified several challenges that the annotators faced. First, we observe that the annotators had issues with the target of evaluation, i.e. the thing or person being evaluated. This is apparent in the already given example (1) above. By looking at the classification choice, the fact that one annotator chose both Affect and Appreciation and marked the expressed meanings as ambiguous indicates that they were unsure and thought that both meanings could apply. In this specific sentence, the annotator was unsure whether the state of being lost is evaluated (which corresponds to Appreciation) or if the speaker is expressing their own feelings about the state (Affect). In the examples (3) and (4) below, both annotators agree that the sentences are evaluative, but we can see an interplay of different Attitude categories being used.

(3) *We also got a bunch of people who, **I guess, just didn’t really like the music or something.*** (classified as A1: Affect and as A2: Judgement).

(4) *I think it’s been a **weird** 100 years.* (classified as A1: Appreciation and as A2: Judgement).

The source of this ambiguity is not paying attention to the right target of evaluation. In (3), one annotator focused on the linguistic item *didn't really like,* which expresses an emotion, but not the one of the speaker. We instructed the annotators to focus on the emotions expressed by the speaker, not those of other people that they mention in their talk. The other annotator focused on the right target of evaluation, which is *a bunch of people* who are being assessed by the speaker. Therefore, it is not Affect that is being expressed, but Judgement. In (4) *weird* is an explicit linguistic item typically associated with Judgement values, which could be a reason why one annotator classified the sentence as Judgement. However, when taking into account the target of evaluation, it becomes clear that inanimate object *years* is being assessed in terms of quality or valuation (Appreciation). Even so, one could argue that explicitly an assessment of the years is expressed and that implicitly, the speaker is offering a Judgement of why the things are the way they are. This illustrates the inherent nature of Appraisal, which can express both explicit and implicit meanings simultaneously and which are not always easy to differentiate (Fuoli, 2018, p. 235), which is precisely the next challenge we identified in the students' annotations.

Second, we find that the annotators struggled to classify explicitly or implicitly expressed evaluative meanings. The degree of subjective interpretation plays a role in which of these meanings is classified in the end, as shown in example (5) below. One annotator classified the sentence as evaluative (Judgement) and the other as Appreciation. By looking at the speaker's choice of the words that colour the sentence with negative prosody, we observe that implicitly, some negative Judgement of the person is being expressed (see Martin & White, 2005, p. 67 for strategies for invoking Judgement). This meaning, which requires a subtle reading of the text and its co-text, is successfully captured by one of the annotators. We believe the other annotator chose Appreciation by identifying the target of evaluation (*occasional miss*) and incorrectly concluding that the speaker is expressing an assessment of an event/thing.

(5) *You just tend to remember the occasional miss, you do.*

Annotating on the sentence level carries the disadvantage of incomplete or wrong classification due to the presence of nested explicit and implicit meanings. This poses problems for identification (what the annotators should focus on in the sentence), but also classification of evaluative meaning (low agreement between annotators occurs if they both focus on a different thing in the sentence). Our guidelines instructed the annotators to focus on the overall evaluative meaning of the sentence, which meant that the students could either focus on the explicit or the implicit, or both. In case multiple meanings were expressed (either explicit or implicit), they were allowed to classify them using multiple Attitude categories. However, we did not tell them to specify or somehow mark which of those meanings they identified were, in fact, explicit or implicit. Therefore, structuring the guidelines in this way contributed to different inter-subjective choices by the annotators, which resulted in low agreements. One way to address this issue is to specify in the annotation guidelines whether the focus of the annotation should be on explicit or implicit evaluative meaning, i.e. linguistic instances used to express it. Future annotation schemes should include this option if they opt for sentence-level Appraisal annotation. As stated before, previous work (Imamovic et al., 2024) operated on the token-level. While human annotators in that study reached higher agreement scores, this was not mirrored in machine annotation using the ChatGPT model. Nevertheless, future research should consider new experiments and prompt designs for token-level annotation to reduce ambiguity due to the level of explicitness.

Next, we find that another problem in the classification of Appraisal is the issue of long sentences, exemplified below in (6). In the given sentence, two explicit expressions of Appraisal are present: an emotion adjective and a stance adverb. Both annotators classified the sentence; however, the disagreement stems from the fact that one annotator classified only the meaning expressed in the first clause (*furious*) and the other classified only the one expressed in the second clause (*absolutely*). The result is that both annotators fail to identify and classify the meaning present in another clause, i.e. both evaluative meanings do not get classified. This shows that one of the meanings does not get annotated simply because the annotator might miss the expression or decide to classify only one of the meanings expressed. Even if both meanings are classified, it is difficult to pinpoint which Appraisal category corresponds to which evaluative expression. The possible solution for this issue is to separate the sentences into clauses and then let them be annotated by annotators.

(6) *Well, the biologists get* ***furious*** *with me for saying that, because we have* ***absolutely*** *no evidence for life beyond Earth yet.* (classified as A1: Affect and as A2: Judgement, Appreciation).

Finally, the low IAA between linguists in training for some domains is due to the lack of experience with annotations, English proficiency and lower understanding of semantics/pragmatics. Other problems in the annotations of linguists in training (those with a better English level) were long sentences, TED talk specialized genre (lack of expertise knowledge on the subject of the TED talk, which also impacts the overall understanding of sentences), annotation of implicit and explicit Appraisal, subjective interpretation, and the target of evaluation. Even those annotators who have C1/C2 English level still show disagreements due to these factors listed. This means that even near-native-like speakers struggle with Appraisal unless delimitations are clearly specified in the annotation guidelines (which ours did not contain). These can be marked as two different "kinds of disagreement", i.e. disagreements that occurred due to English proficiency and low pragmatic skills and those disagreements that occurred because of practical reasons that the annotators with better English skills struggled with. The question remains, then – how do LLMs perform in the identification and classification of Appraisal, when compared to the gold standard?

## 5 Automatic Annotation

In this section, we report on the automatic annotation of the classes using LLMs. First, we report on the development and comparison of three prompts. Following this, using the best-performing prompt, we contrast the performance of three LLMs. The best-performing prompt is finetuned with QLora. The LLM selection is explained in Appendix 6. All models mentioned were used via the SAIA API. The assistance of ChatGPT 5.0. was used to perform preliminary data coding for the creation of graphs. All content generated by the tool was checked and approved by the authors. All code and output data underlying this article are available on GitHub (https://github.com/happy522/Challenges-in-annotations-by-humans-and-LLMs). All original videos of Emotionaliz*TED* are publicly available on the TED website: ted.com/, lastly accessed by the authors 01/15/2026. The Emotionaliz*TED* corpus will be made available within the realm of another publication that is currently in press.

### 5.1 Prompt Design

We compare the performances of three prompts by testing them on a subsample of the data, using the annotations of the senior researcher as gold labels. The aim was to identify the best-performing prompt, as the choice of a proper prompting scheme is crucial (Cruickshank et al., 2025). For automatic annotation, we follow the same procedural steps as those used in manual annotation. First, the sentences are annotated as either evaluative or not. The evaluativeness of the sentence is translated to a binary classification problem, with the sentence being either evaluative (1) or not (0). In a second step, the sentences are allocated to one of three Appraisal categories (subclasses), but only if they have been labelled as evaluative before. Because of this two-step process, the task can be classified as a multi-turn dialogue. The Appraisal categories constitute the classes for the multiclass classification. In addition, we introduce the Ambiguous class. Just like humans, the model was allowed to select multiple classes if the sentence was ambiguous. In the course of this paper, we do not further report on the Ambiguous class. No overlap was observed in the sentences classified as ambiguous; consequently, all measures equal zero.

| | Version 1 | Version 2 | Version 3 |
|---|---|---|---|
| Common characteristics | Multi-turn dialogue<br>Context<br>Persona description<br>Constraints to prevent hallucination | | |
| Distinct characteristics | Few-shot prompting<br>Chain-of-thought prompting<br>Avoidance of ambiguity | Zero-shot prompting<br>Less detail<br>Class description is shorter | Few-shot prompting<br>Rather than real examples, we provide indicative adjectives and n-grams |

**Table 2** Overview of prompt variations that were tested for the multiclass classification. The characteristics of all three prompt versions have in common are that they offer context, a persona description, constraints to prevent hallucination. Moreover, we worked in a multi-turn dialogue scenario. Version 1 and 3 are few-shot prompting approaches, while version 2 is a zero-shot prompting approach. Next to this, less detail and shorter class descriptions are provided in version 2. In version 3, we provide indicative adjectives and n-grams instead of real examples. For version 1 prompt, ambiguity is avoided. Next to this, version 1 is a Chain-of-thought prompting approach

For testing all prompts, the model qwen3-30b-a3b-instruct-2507[5] (Qwen) is employed, as it performs particularly well on reasoning tasks (Qwen Team, 2025). All models are accessed via the GWDG API[6], hosted by SAIA. First, we design the prompt for the identification of the evaluative class (binary classification). For this task, we read an F1-Score = 0.79 using our initial prompt, which is displayed in Appendix 2. Given this strong initial performance, the subsequent design and comparison of three prompts focus exclusively on identifying the subclasses, i.e. Appraisal categories. All three prompt versions can be seen in the Appendices III-V. An overview of differences and common elements in the prompts is given in Table 2.

We craft the first version of the prompt (Appendix 3), building on the learnings from Hamilton et al. (2024), Hou et al. (2024) and Cruickshank et al. (2025), who worked with complex annotation schemes in the fields of linguistics or social science coding. In the end, the prompt involves a persona description and chain-of-thought prompting, as it elicits reasoning (Hou et al., 2024). Chain-of-thought prompting instructs the model to provide reasoning, providing an interpretable window into the behaviour of the model (Wei et al., 2022; Hou et al., 2024). It has consistently improved the output of LLMs and helped to prevent hallucination (Wei et al., 2022; Chen et al., 2023; Cruickshank et al., 2025). We avoid ambiguity in the wording of the instruction (no use of demonstrative pronouns). Like in all prompt versions, context is provided by including a reference to the previous and subsequent text. Additionally, we ask the model to output a probability for each prediction, with a float between 0.0 and 1.0. In addition, the model must generate a list of three tokens from the target text that support the model's decision. For the second version of the prompt (see Apx. IV), we also include a persona description and choose a few-shot design. While we do give examples, they are not real examples from the corpora but rather indicative adjectives or n-grams. For the third prompt version (see Apx. V), our approach was to follow the linguistic theory very closely. We keep the context shorter, given through the explanation of the classes and provide less detail. Still, the prompt remains similar to the others as it is a multi-turn prompt, a few-shot approach, context, and a persona description. In all three prompts, we try to prevent hallucinations by giving constraints like "Do NOT invent extra information [...]". We find that version 3 outperforms the other prompts with an F1-score of 0.49 (vs. version 1 reaching an F1-score = 0.43 and version 2: F1-score = 0.24). Although the scores are relatively low, likely due to class imbalance in the subsample, the results suggest that prompt version 3 consistently outperforms the alternatives. All data used to test the prompt versions were excluded from further analyses.

[5] https://huggingface.co/Qwen/Qwen3-30B-A3B-Instruct-2507

[6] docs.hpc.gwdg.de/index.html

### 5.2 Results of the Automatic Annotation

Overall, all three models reach the best F1-score for the binary classification task, with F1-score >= 0.68 (see Table 3). Performance drops when it comes to the identification of the Appraisal categories Affect, Judgement, and Appreciation (see Table 4). In both tasks, Qwen outperforms the Llama and the Mistral models. Next to this, all three models reach higher precision values than recall in both tasks. This reflects the prompt, where conservative decision-making was encouraged. Also, this decision is mirrored in the high precision values, which reveal that the model predicts few false positives but misses actual positives. Nevertheless, conservative decision-making was necessary to lower the chance of hallucination.

| Model | Precision (macro) | Recall (macro) | F1-score (macro) |
|---|---|---|---|
| Llama-3.3-70b-instruct | 0.71 | 0.70 | 0.65 |
| Qwen3-30b-a3b-instruct-2507 | 0.70 | 0.70 | 0.68 |
| Mistral-large-instruct | 0.72 | 0.70 | 0.67 |

**Table 3** Results for the binary classification (evaluativeness): Ø macro precision, recall and F1-score across models. Precision values are slightly higher than recall for Llama (0.71 vs 0.70) and Mistral (0.72 vs 0.70). For Qwen, both precision and recall equal 0.70. Qwen reaches the highest overall macro F1-score with 0.68

| Model | Precision (macro) | Recall (macro) | F1-score (macro) |
|---|---|---|---|
| Llama-3.3-70b-instruct | 0.87 | 0.25 | 0.33 |
| Qwen3-30b-a3b-instruct-2507 | 0.78 | 0.31 | 0.40 |
| Mistral-large-instruct | 0.85 | 0.22 | 0.29 |

**Table 4** Results for the multilabel classification (Appraisal categories): Ø macro precision, recall and F1-score across models. All models demonstrate high precision values, while recall is low (Mistral: 0.22, Qwen: 0.31, Llama: 0.25)

The three LLMs evaluated show broadly similar behaviour but with a consistent bias toward predicting Affect and clear difficulty distinguishing Judgement. In Fig. 6 and Fig. 7, the per-class results per model are displayed. On the binary evaluative task (model vs. gold), Qwen achieved the highest accuracy (68.08%), followed by Mistral (66.07%) and Llama (65.18%). Qwen also posts the best weighted and macro averages (weighted F1-score = 0.68, macro F1-score = 0.68), indicating a small but consistent overall edge.

In the multiclass Appraisal task, Qwen again ranks highest by accuracy (61.05%) and by macro F1-score = 0.52, with Mistral (accuracy 60.57%, macro F1 = 0.492) and Llama (accuracy 59.86%, macro F1-score = 0.479) trailing. A consistent pattern emerges in the per-class results: Affect is identified reliably (high recall: 0.84-0.90 and F1-score = 0.73 - 0.74 across models), Judgement is the weakest target (very low recall 0.16–0.22 and F1-score = 0.26 - 0.32), and Appreciation sits in the middle (with recall ranging between 0.38-0.54 and F1-score ranging between 0.42-0.49 ). The confusion matrices confirm that many true Judgement and Appreciation instances are misclassified as Affect (see Fig. 8), which explains the inflated Affect recall at the expense of other classes. Similarly, binary confusion matrices show higher false negatives for the positive class (e.g. Qwen: 109 false negatives of class 1). These results suggest model outputs are dominated by a safer Affect prediction and that class imbalance or label ambiguity is likely contributing to systematic misclassification.

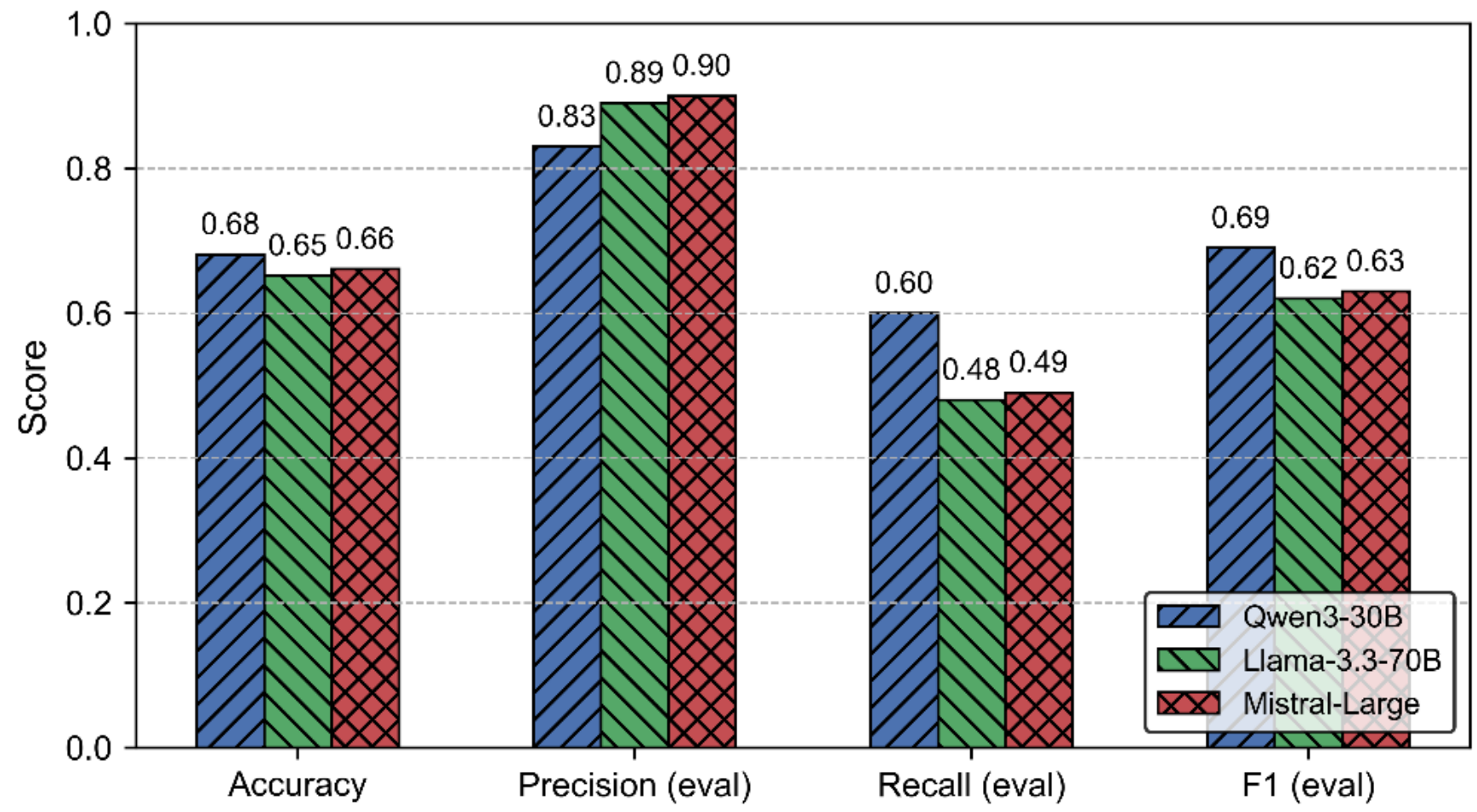

**Fig. 6** LLM classification results for the evaluative class. Again, precision values are higher than recall. Qwen demonstrates the highest F1-score with F1=0.69, while Llama and Mistral reach the highest precision values (Llama: 0.89, Mistral: 0.90)

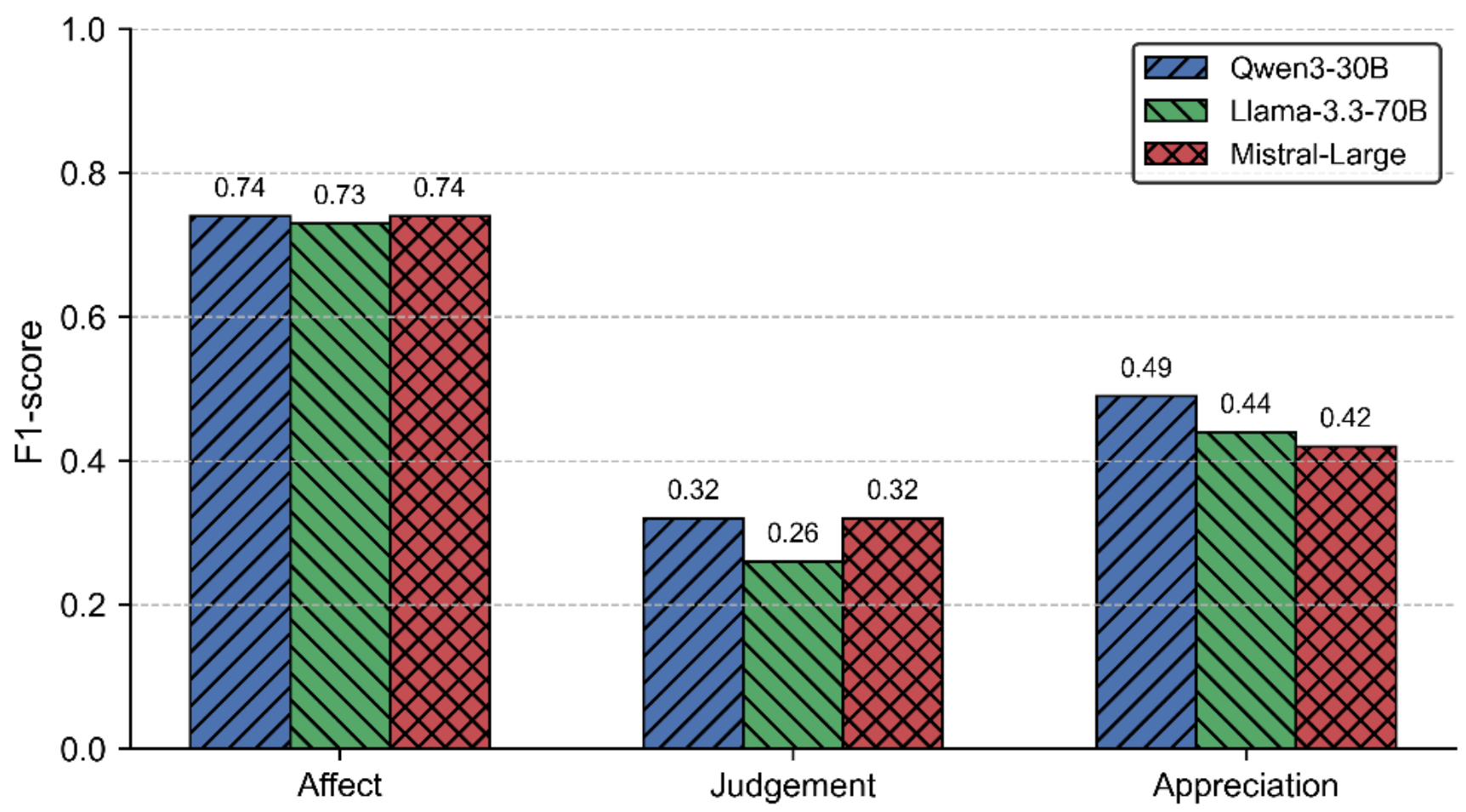

**Fig. 7** LLM classification results for the subclasses Affect, Judgement, and Appreciation. The F1-scores for Affect are >= 0.73, while Appreciation F1 >= 0.42. F1 for Judgement ranges between 0.26 and 0.32

Next to precision, accuracy, recall, and F1-score, we also calculated the IAA using Krippendorff's alpha for the per-class ratings of the senior researcher and the models (see Fig. 8). Agreement scores range between fair and moderate. The highest agreement the models and the senior researcher achieve is in the identification of the Affect class. Qwen reaches the highest agreement in the evaluative and the Appreciation class, while both Llama and Qwen reach only low agreement in the Judgement class.

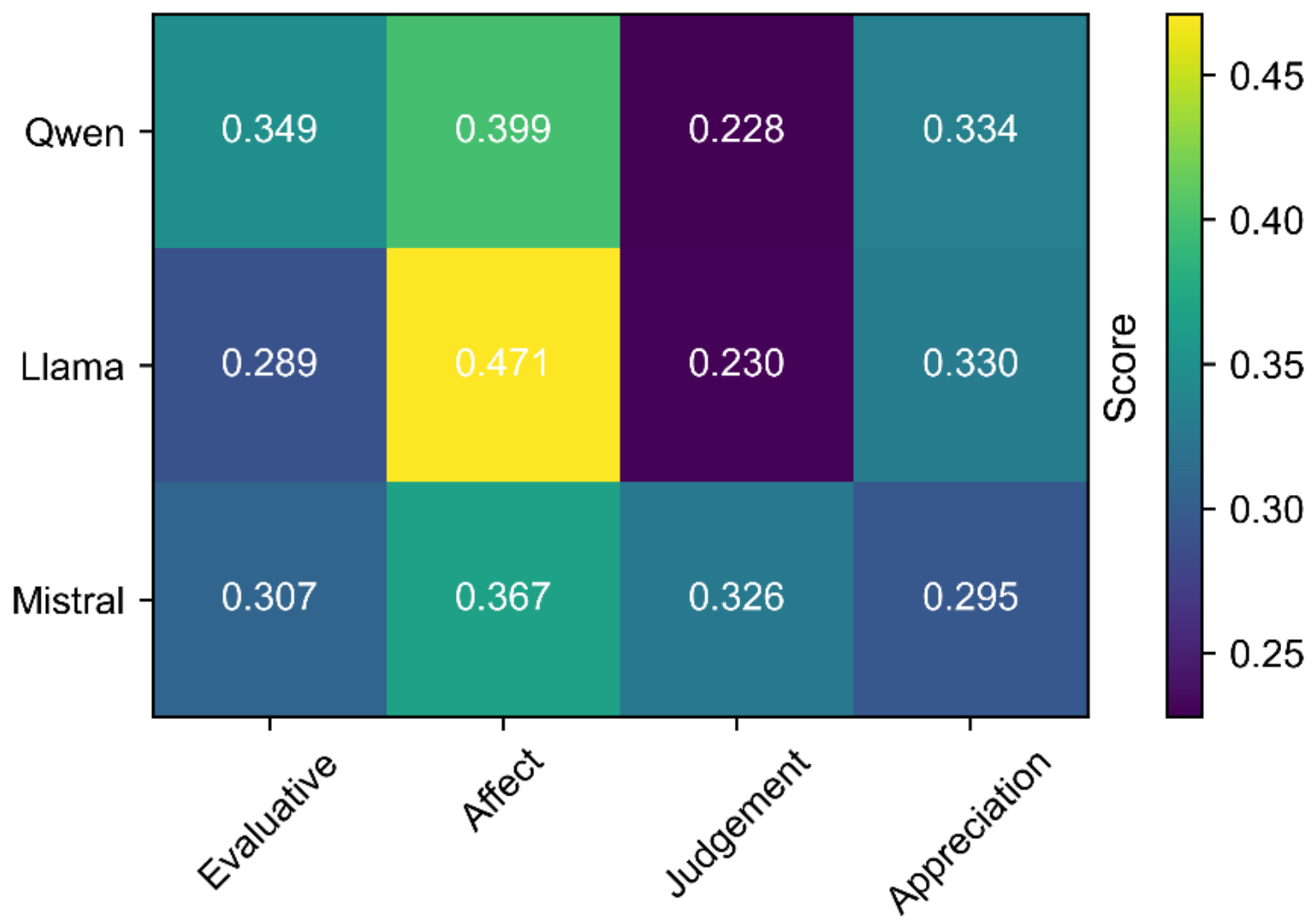

**Fig. 8** Heatmap for the agreement between the trained linguist and LLMs using Krippendorff's alpha. Agreement scores are generally fair or moderate. The highest agreement is between Affect and Llama (α = 0.471) and Affect and the other two models (Qwen: α = 0.399, Mistral: α = 0.367). IAA is lowest for the Judgement class for Qwen (α = 0.228) and Llama (α = 0.23). For the Evaluative and Appreciation class, IAAs between α = 0.289 and α = 0.349 are obtained

### 5.3 Finetuning using QLora

We conducted two main fine-tuning experiments using Qwen. First, the model was fine-tuned for sentence-level binary classification on the gold standard with a 60/40 train-test split. Second, Qwen was fine-tuned for the identification of the Appraisal categories, restricting training and evaluation to sentences identified as evaluative in the binary stage, thereby forming a staged pipeline aligned with Appraisal theory. In both experiments, fine-tuning was performed using QLoRA with 4-bit NF4 quantization and low-rank adapters, and supervision was applied only to the generated label tokens while surrounding sentences were used strictly as contextual cues.

The fine-tuned binary classifier attains ~0.70 accuracy with strong sensitivity to evaluative content (Evaluative recall = 0.87, Evaluative F1-score = 0.77), modestly improving the model's ability to capture evaluative sentences compared with direct prompting (direct prompting accuracy = 0.68, Evaluative F1-score = 0.69). By contrast, as Fig. 9 shows, the multi-label Appraisal model shows substantially weaker performance: low precision across categories, micro/ macro F1-score = 0.34 / 0.30, and very low exact-match subset accuracy (~0.04), reflecting difficulty in reliably predicting overlapping Appraisal categories despite high recall for some labels. Compared to direct multiclass prompting (which achieved a higher macro F1-score = 0.52 and accuracy = 0.61 under mutually exclusive assumptions), our fine-tuned multi-label approach is more sensitive but less precise, highlighting that fine-tuning substantially benefits coarse evaluative detection while fine-grained, multi-label Appraisal recognition remains an open challenge likely driven by label imbalance, co-occurrence, and the inherent nuance of Appraisal categories.

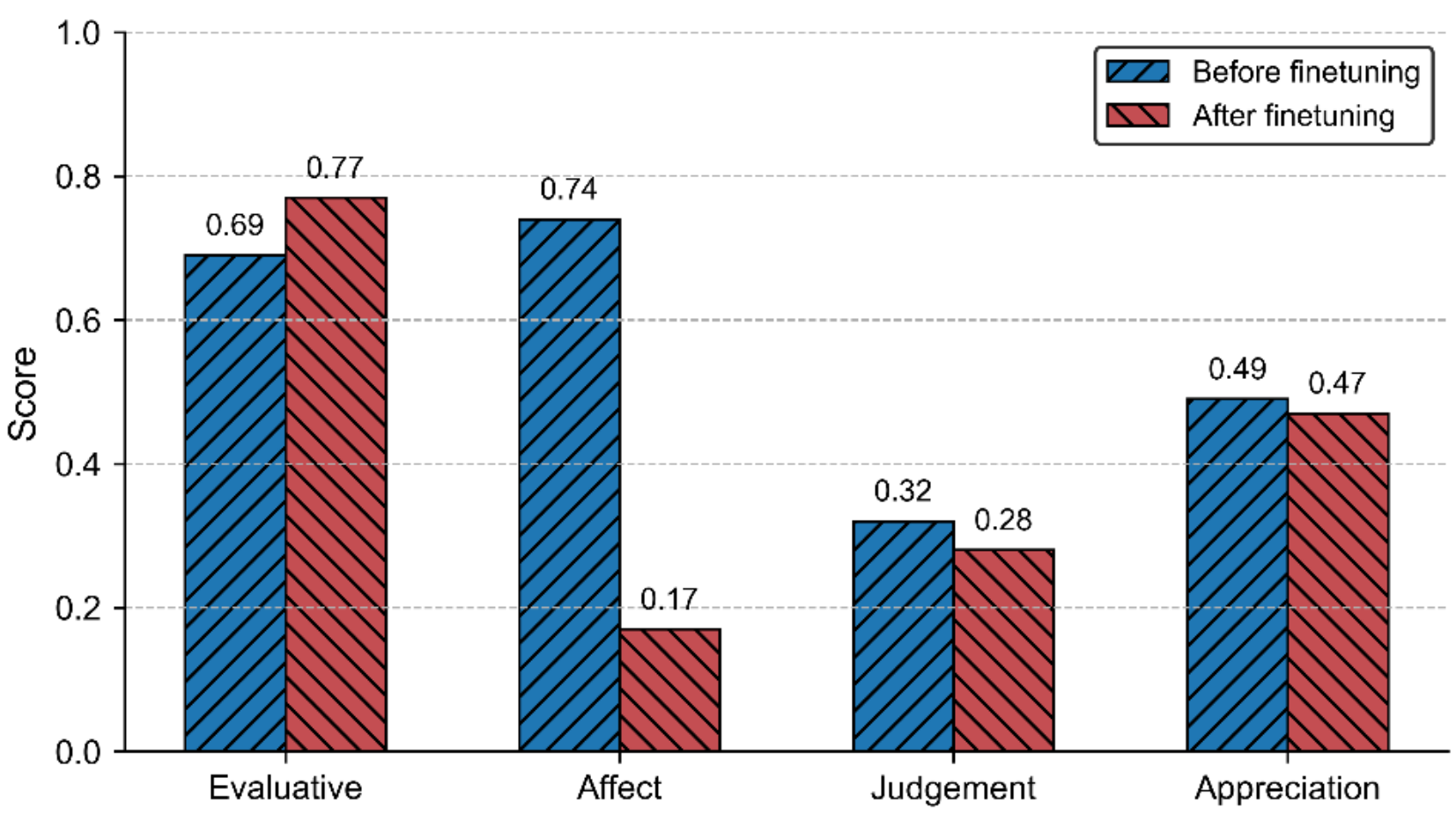

**Fig. 9** Performance change of the Qwen model before and after finetuning looking at F1-score. The graph shows that while performance increases for the detection of the evaluativeness of the text (+0.8), performance drops for the classification of the subclasses Affect, Judgement, and Appreciation. The contrast is especially stark for the Affect class, where the F1-score drops from F1=0.74 to F1=0.17. Performance drops slightly for the Judgement class (-0.4) and Appreciation (-0.2)

## 6 Analysis: Comparing Manual and Automatic Labelling

In this section, we compare the results of manual and automatic labelling. For this purpose, we calculated the average agreement using Krippendorff's alpha per class, as demonstrated in Table 5. We do not conduct domain-wise interpretation of the results, as this is beyond the scope of this paper. Instead, we focus on agreement between humans and LLMs.

| | Evaluative | Affect | Judgement | Appreciation |
|---|---|---|---|---|
| Linguists in training | 0.38 | 0.23 | 0.19 | 0.20 |
| Linguists in training and senior researcher | 0.30 | 0.25 | 0.16 | 0.25 |
| Qwen and senior researcher | 0.35 | 0.40 | 0.23 | 0.33 |
| Llama and senior researcher | 0.30 | 0.47 | 0.23 | 0.33 |
| Mistral and senior researcher | 0.30 | 0.37 | 0.33 | 0.30 |

**Table 5** Average Krippendorff's alpha across domains per class

With a fair agreement of $\alpha = 0.38$, humans reach the highest agreement across classes in detecting the evaluativeness of the text. Interestingly, agreement drops when measured between the senior researcher and the linguists in training ($\alpha = 0.30$). Next to this, humans struggle most with the identification of Judgement, reaching only poor agreement with $\alpha <= 0.19$. Interestingly, machines obtain fair agreement with the senior researcher for this class. The agreement between Mistral and the senior researcher is highest for this class, the only time when Mistral outperforms the other models. Humans reach the second-highest agreement in the identification of Affect (fair agreement). Agreement between the senior researcher and Qwen and Llama is highest across all classes and constellations, with the models reaching moderate agreement scores. In the detection of Appreciation, both humans and machines reach fair agreement, with human agreement ranging from $\alpha = 0.20$ - $0.25$ and agreement between the machines and the senior researcher lying at $\alpha >= 0.30$. Concluding this analysis, humans agree most in the identification of the evaluativeness, while machines obtain the highest agreement in the identification of Affect. Notably, machines reach a higher agreement with the senior researcher than students in the identification of all Appraisal categories. Interestingly, Llama on average reaches higher agreement scores than Qwen, although Qwen outperformed the other models measured with traditional evaluation metrics applied in NLP, precision, recall, accuracy, and F1-score.

## 7 Conclusion

In this paper, we explore the performance of humans and LLMs for the annotation of evaluative language. On the one hand, we focus on the problems human annotators face when dealing with such a complex linguistic phenomenon. On the other hand, we test the feasibility of LLMs for this complex task. We are also interested if humans and LLMs face similar problems in the annotation task.

Addressing **RQ1**, we found that certain classes of Attitude seem to be more challenging for annotators. However, this result is dependent on the domain a TED talk belongs to. Thus, human annotators perform well (as measured against the gold standard by the senior linguist) on Affect categories in Business and Psychology, on Appreciation also in Psychology, but in Entertainment as well, although the results are, in general, worse than for Affect. For the category of Judgement, they reach a threshold of 0.50 in Arts only, showing the overall worst result in annotation.

Now, answering **RQ2**, we find that LLMs can meet the expectations expressed in the scientific community to some extent. Our results demonstrate that LLMs can aid in complex annotation tasks, serving as data annotators. In this case study, results show that LLMs exceed the performance of non-expert annotators in the labelling of some classes. Specifically, LLMs reach higher agreement with senior researcher annotations than linguists in training for the Appraisal categories. In the detection of the evaluativeness of a sentence, they come close to human performance.

In terms of **RQ3**, we see that LLMs struggle with the same Appraisal categories as humans (Judgement and Appreciation). Nevertheless, both humans and LLMs show moderate agreement at best in the detection of all classes. For LLMs, this is true, although prompts have been crafted and tested carefully. This indicates the complexity and subjectivity of the task, enabling two conclusions. First, LLMs for automatic annotation are not a panacea for complex annotation task resolution. Second, the implementation of LLMs for label generation is still an extensive process, calling for an interdisciplinary team of experts.

**8 Discussion**

Our results show that the overall agreement rates are still too low to consider either human- or machine-generated annotations reliable enough for training supervised classification models. This finding suggests that, although the use of LLMs for annotation purposes shows potential, they should not be regarded as a panacea. Our study shows that the difficulties inherent in manual annotation are not necessarily resolved by LLMs. Consequently, for every new task, it remains essential to verify whether an LLM actually outperforms non-expert annotators who have received training. Based on the results of this study, we recommend case-by-case validation for LLM-generated annotations, especially in highly specialized domains such as Appraisal theory and for subjective tasks. . In our case, such validation shows that LLM consistently surpasses the accuracy of linguists in training and attains a level of performance comparable to that of a senior researcher, if the research design is conceptualized carefully and evaluatedHowever, while generating annotations with LLMs is technically feasible and less cost- and time-expensive, the process is far from simple and presently requires an interdisciplinary research team. Likely, a future in which individual linguists without specific technical skills can independently produce high-quality LLM-based annotations based on prompting alone is out of reach.

While we agree with Ziems et al. (2024) that LLMs can be employed to aid in annotation workflows, we highlight the importance of a high-quality benchmark against which the machine-generated output is to be evaluated. To do so, we need a gold-standard, i.e. manually labelled corpus for comparison. This is especially challenging for subjective tasks. Future research should consider a perspectivist approach for subjective tasks (see, for instance, Romberg et al. 2025). A question appearing in this context is the principle of a ground truth, where one single gold label per instance is assumed to be true (Plank, 2022). On the one hand, given the ambiguity and subjectivity of Appraisal, we think that several answers could be plausible, with genuine human label variation following Cabitza et al. (2023). The authors took several precautions advertised in the literature to avoid error in the labelling process: intensive training of annotators over the course of a semester, with homework (Gadiraju et al., 2015) and collective discussion (Muller et al., 2021). Still, we consider some aspects in the design of the qualitative study as reasons for low agreement. First, we find that the document level of annotation could be selected differently. While annotators operated on a sentence level in this study, our evidence suggests that a span level of annotation could be a more fruitful approach. In this study, we employed a sentence-level approach instead of a token- or span-level one. Our qualitative analysis reveals that there is still a need for specification of an evaluative expression which triggers the evaluative meaning. Thus, it would be better to ask annotators to mark the specific evaluative expression that underlies the label they give to a sentence.

This is also helpful for finding out whether they focused on the explicitly or implicitly expressed attitude, which facilitates a better understanding of whether the annotation result is correct. Thus, we see differentiation between explicitness and implicitness as a second major problem. We struggled to evaluate the assigned labels, since we did not know what the annotators focused on. We also know from other studies (Fuoli et al., 2022) that the lowest agreement is normally achieved for segments containing implicit Attitude, which the authors attributed to the inherently subjective and context-dependent nature of such evaluations. Without knowing if annotators see a sentence as evaluative because of explicit or implicit Attitude, we are unfortunately not able to differentiate this. Our qualitative analysis shows that the annotation guidelines must explicitly state whether annotators should target implicit, explicit, or both types of Appraisal. Annotating whole sentences without this delimitation leads to lower IAA, a problem we briefly discuss in the error-analysis section. Clear guideline specifications are therefore important for reliable annotation.

Another issue in this study is the fact that all human annotators involved in our study are non-native speakers of English. At the same time, we believe that it is possible to compare their performance to LLMs, since LLMs are trained with data which was not entirely produced by native speakers of English. Besides that, many students do have a high level of proficiency in English, and our senior researcher has C2 level, which means a professional and native-like user according to CEFR[7].

**References**

Aikhenvald, A. Y. (2004). *Evidentiality.* Oxford: Oxford University Press.

Alsina, V., Espunya, A., & Naro, M. W. (2017). An appraisal theory approach to point of view in Mansfield Park and its translations. *International Journal of Literary Linguistics*, *6*(1), 1–28. https://doi.org/10.15462/ijll.v6i1.103

Bednarek, M. (2006). *Evaluation in Media Discourse: Analysis of a Newspaper Corpus.* London: Continuum.

Bednarek, M. (2008). Analyzing language and emotion. In M. Bednarek (Ed.), *Emotion talk across corpora*, pp. 1–26. London: Palgrave Macmillan.

Bednarek, M. (2009). Language patterns and ATTITUDE. *Functions of language, 16*(2), 165–192.

Biber, D., & Finegan, E. (1988). Adverbial stance types in English. *Discourse processes, 11*(1), 1–34.

Biber, D., & Finegan, E. (1989). Styles of stance in English: Lexical and grammatical marking of evidentiality and affect. *Text-interdisciplinary journal for the study of discourse, 9*(1), 93–124.

Bloom, K. (2011). *Sentiment analysis based on appraisal theory and functional local grammars*. PhD Thesis. Illinois Institute of Technology.

Busetto, N., & Delmonte, R. (2019). Annotating Shakespeare's Sonnets with Appraisal Theory to Detect Irony. In *Proceedings of the Sixth Italian Conference on Computational Linguistics (CLiC-it 2019)*, pages 55–62, Bari, Italy. CEUR Workshop Proceedings.

Cabitza, F., Campagner, A., & Basile, V. (2023). Toward a perspectivist turn in ground truthing for predictive computing. In *Proceedings of the AAAI Conference on Artificial Intelligence, 37*(6), 6860–6868.

Cavasso, L., & Taboada, M. (2021). A corpus analysis of online news comments using the Appraisal framework. *Journal of Corpora and Discourse Studies, 4*, 1–38. http://doi.org/10.18573/jcads.61

Chen, J., Chen, L. Huang, H., & Zhou, T. (2023). When do you need Chain-of-Thought Prompting for ChatGPT? In *arXiv preprint, arXiv:* https://arxiv.org/abs/2304.03262

Cohen, J. (1960). A coefficient of agreement for nominal scales. *Educational and Psychological Measurement, 20,* 37–46. https://doi.org/10.1177/001316446002000104

Conrad, S., & Biber, D. (2000). Adverbial marking of stance in speech and writing. In S. Hunston & G. Thompson (Eds), *Evaluation in Text: Authorial Stance and the Construction of Discourse*, pp. 56–73. Oxford: Oxford University Press.

Cruickshank, I. J., & Ng, L. H. X. (2025). Prompting and Fine-Tuning Open-Sourced Large Language Models for Stance Classification. In *Association for Computing Machinery.* https://doi.org/10.1145/372581

[7] Common European Framework of Reference for Languages

Diewald, G., & Smirnova, E. (2010). *Linguistic realization of evidentiality in European languages*, *49.* Berlin: De Gruyter Mouton. https://doi.org/10.1515/9783110223972

Dong, M., & Fang, A. (2023). Appraisal theory and the annotation of speaker-writer engagement. In *Proceedings of the 19th Joint ACL-ISO Workshop on Interoperable Semantics (ISA-19)*, pp. 18–26, Nancy, France. Association for Computational Linguistics.

Eggins, S., & Slade, D. (1997). *Analysing casual conversation*. London: Cassell.

Fuoli, M. (2018). A stepwise method for annotating APPRAISAL. *Functions of language, 25*(2), 229–258.

Fuoli, M., Littlemore, J., & Turner, S. (2022). Sunken ships and screaming banshees: Metaphor and evaluation in film reviews. *English Language & Linguistics, 26*(1), 75–103.

Gadiraju, U., Fetahu, B., & Kawase, R. (2015). Training workers for improving performance in crowdsourcing microtasks. In *10th European Conference on Technology Enhanced Learning,* Toledo, Spain, pp. 100–114. https://doi.org/10.1007/978-3-319-24258-3

Glynn, D., & Sjölin, M. (2014). *Subjectivity and epistemicity: Corpus, discourse, and literary approaches to stance*. Lund: Lund University.

Halliday, M. A. K. (2004/1994). *An Introduction to Functional Grammar.* London: Edward Arnold. (2004 third edition revised by C. M. I. M. Matthiessen).

Halliday, M. A. K., & Matthiessen, C. (2014). Halliday's introduction to functional grammar (4th ed.). London: Routledge.

Hamilton, K., Longo, L., & Bozic, B. (2024). GPT Assisted Annotation of Rhetorical and Linguistic Features for Interpretable Propaganda Technique Detection in News Text. In *Companion Proceedings of the ACM Web Conference 2024*, Singapore, Singapore, pp. 1431–440. Association for Computing Machinery. https://doi.org/10.1145/3589335.365190.

Hou, C., Zhu, G., Zheng, J., Zhang, L., Huang, X., & Zhong, T. (2024). Prompt-based and fine-tuned GPT models for context-dependent and- independent deductive coding in social annotation. In *Proceedings of the 14th learning analytics and knowledge conference*, Kyoto, Japan, pp. 518–528.

Hunston, S., & Thompson, G. (2000). *Evaluation in text: Authorial stance and the construction of discourse: Authorial stance and the construction of discourse*. Oxford: Oxford University Press.

Hunston, S. (2010). *Corpus approaches to evaluation: Phraseology and evaluative language*. New York: Routledge.

Imamovic, M., Deilen, S., Glynn, D., & Lapshinova-Koltunski, E. (2024). Using ChatGPT for Annotation of Attitude within the Appraisal Theory: Lessons Learned. In *Proceedings of the 18th Linguistic Annotation Workshop (LAW-XVIII)*, pages 112–123, St. Julians, Malta. Association for Computational Linguistics.

Imamovic, M., Deilen, S., Glynn, D., & Lapshinova-Koltunski, E. (2025). Appraisal in translation: Evaluative expressions in German transcripts of English TED Talks. *SKASE Journal of Translation and Interpretation, 18*(2), 149–164. https://doi.org/10.33542/JTI2025-S-8

Imamovic, M. (2026). Humans versus LLMs: Optimising Prompting Strategies for ChatGPT and Gemini in APPRAISAL Annotation of English TED Talks. In M. L. Carrió-Pastor & C. Periñán-Pascual (Eds). *Research Trends in Discourse Analysis and Language Acquisition in the Artificial Intelligence Era.* Lausanne, Switzerland: Peter Lang Verlag. https://doi.org/10.3726/b23335

Kärkkäinen, E. (2006). Stance taking in conversation: From subjectivity to intersubjectivity. *Text & Talk, 26,* 699–731.

Knierim, A., & Heid, U. (2025). Argumentation in political empowerment on Instagram. In *Proceedings of the 9th Joint SIGHUM Workshop on Computational Linguistics for Cultural Heritage, Social Sciences, Humanities and Literature (LaTeCH-CLfL 2025),* ed. by Anna Kazantseva et al. Albuquerque, New Mexico. Association for Computational Linguistics, pp. 97–108. https://aclanthology.org/2025.latechclfl-1.10/

Krippendorff, K. (2011). Computing Krippendorff's alpha-reliability. https://api.semanticscholar.org/CorpusID:59901023

Langacker, R. (1985). Observations and speculations on subjectivity. In J. Haiman (Ed.), Iconicity in Syntax, 109–150. Amsterdam: John Benjamins. https://doi.org/10.1075/tsl.6.07lan

Lyons, J. (1977). Semantics. Cambridge: Cambridge University Press.

Martin, J. R. (1992). English text: System and structure. Amsterdam: Benjamins.

Martin, J. R., & White, P. R. (2005). The Language of Evaluation. New York: Palgrave Macmillan.

Muller, M., Wolf, C. T., Andres, J., Desmond, M., Joshi, N. N., Ashktorab, Z., Sharma, A., Brimijoin, K., Pan, Q., & Duesterwald, E. (2021). Designing Ground Truth and the Social Life of Labels. In Proceedings of the 2021 CHI Conference on Human Factors in Computing Systems, pp. 1–16.

Munday, J. (2012). *Evaluation in translation: Critical points of translator decision-making*. London: Routledge.

Palmer, F. R. (2014). Modality and the English modals. London: Routledge.

Pang, B., & Lee, L. (2008). Opinion Mining and Sentiment Analysis. *Foundations and Trends® in Information Retrieval, 2,* pp. 1–135. https://doi.org/10.1561/1500000011

Pangakis, N., Wolken, S., & Fasching, N. (2023). Automated annotation with generative AI requires validation. In arXiv preprint, arXiv: https://arxiv.org/abs/2306.00176

Read, J., & Carroll, J. (2012). Annotating expressions of Appraisal in English. *Language Resources and Evaluation, 46*(3), 421–447.

Romberg, J., Maurer, M., Wachsmuth, H., & Lapesa, G. (2025). Towards a perspectivist turn in argument quality assessment. In *Proceedings of the 2025 Conference of the Nations of the Americas Chapter of the Association for Computational Linguistics: Human Language Technologies (Volume 1: Long Papers)*, 7458–7485.

Stingo, M., & Delmonte, R. (2016). Annotating satire in Italian political commentaries with appraisal theory. In *NLPMJ-Natural Language Processing meets Journalism*, pp. 74–79.

Taboada, M., & Grieve, J. (2004). Analyzing appraisal automatically. In *Spring symposium on exploring attitude and affect in text,* American Association for Artificial Intelligence, Stanford, AAAI Technical Report SS-04-07.

Taboada, M., & Carretero, M. (2012). Contrastive analyses of evaluation in text. *Linguistics and the Human Sciences 6(*1-3), 275–295.

Taboada, M., Carretero, M., & Hinnell, J. (2014). Loving and hating the movies in English, German and Spanish. *Languages in Contrast, 14*(1), 127–161.

Team, Qwen. (2025). Qwen3 Technical Report. arXiv: https://arxiv.org/abs/2505.09388

Thompson, G., & Alba-Juez, L. (2014). Evaluation in Context. Amsterdam and Philadelphia: John Benjamins.

Toprak, C., Jakob, & N., Gurevych I. (2010). Sentence and Expression Level Annotation of Opinions in User-Generated Discourse. In Proceedings of the 48th Annual Meeting of the Association for Computational Linguistics, pages 575–584, Uppsala, Sweden. Association for Computational Linguistics.

Wei, J., Wang, X., Schuurmans, D., Bosma, B., Ichter, B., Xia, F., Chi, E., Le, Q., & Zhou, D. (2022). Chain-of-thought prompting elicits reasoning in large language models. *In Proceedings of the 36th International Conference on Neural Information Processing Systems. NIPS '22.* New Orleans, LA, USA, pp. 24824–24837.

Yi, Z., Ouyang. J., Xu, Z., Liu, Y., Liao, T., Luo, H., & Shen, Y. (2025). A Survey on Recent Advances in LLM-Based Multi-turn Dialogue Systems." In *ACM Computing Surveys*, *58*(6), pp 1–38. https://doi.org/10.1145/3771090

Yu, D., Li, L., Su, H., & Fuoli, M. (2024). Assessing the potential of LLM-assisted annotation for corpus-based pragmatics and discourse analysis: The case of apologies. *International Journal of Corpus Linguistics, 29*(4), 534–561. https://doi.org/10.1075/ijcl.23087.yu

Zeng, J., Dong, M., & Fang, A. C. (2024). Annotating Evaluative Language: Challenges and Solutions in Applying Appraisal Theory. In *Proceedings of the 20th Joint ACL-ISO Workshop on Interoperable Semantic Annotation@ LREC-COLING 2024*, pages 144–151, Torino, Italia. ELRA and ICCL.

Ziems, C., Held, W., Shaikh, O., Chen, J., Zhangm Z., & Yang, D. (2024). Can Large Language Models Transform Computational Social Science?. *Computational Linguistics, 50*(1), 237–291.

**Appendix 1 Annotation guidelines**

The following table describes the guidelines that the annotators received.

| Column | Explanation/What to do |
|---|---|
| **Evaluative** | Put 1 if the sentence expresses an emotion, judgement (ethics) or aesthetic evaluation. Put 0 if the sentence is purely factual or neutral. If you put 0, leave all other columns blank. |
| **Affect** | Put 1 if the sentence expresses **feelings or emotions** (e.g. *happy, scared, grateful*). |
| **Judgement** | Put 1 if it evaluates **behaviour or character** (e.g. *kind, unfair, brave*). |
| **Appreciation** | Put 1 if it evaluates **things, performance, or phenomena** (e.g. *beautiful, boring, excellent*). |
| **Ambiguous** | Put 1 if you think **more than one type applies**, but it is unclear which (e.g. not sure if it's Affect or Judgement). You must also mark the types you're considering (e.g. Affect = 1, Judgement = 1). |
| **Uncertain** | Put 1 only if you're **not sure if your choice is correct.** If you use this, **please write a short note.** You can use Uncertain if you are not sure that 1) the sentence is evaluative at all, 2) that it expresses Affect, Judgement or Appreciation (e.g. you marked Affect = 1, but you're not confident about it), 3) if the sentence is ambiguous, 4) if the sentence expresses multiple Attitudes (e.g. Affect and Judgement). |
| **Notes** | Use only when you select "Uncertain." Explain briefly what you're unsure about (e.g. “Not sure if it's emotion or factual description”). Use the reasons 1-4 under Uncertain to explain why. |

**Table 6** Annotation instructions for labelling Attitude in text

**Appendix 2 Prompt for binary classification**

You are an annotation assistant for Appraisal Theory (Martin and White, 2005).
Task:
Decide whether the TARGET text contains any evaluative content.

You are given:
- PREVIOUS: text from the row above (may be empty)
- TARGET: the current text to label
- NEXT: text from the row below (may be empty)

Use PREVIOUS and NEXT only as context (e.g. to resolve pronouns or topics).
The label must refer ONLY to the TARGET text.

Coding rule “Evaluative” (binary):
- Evaluative = 1
  -> If the TARGET text expresses an emotion, a judgment of people/behaviour, or an aesthetic/quality/value evaluation of things, performances or phenomena.
- Evaluative = 0
  -> If the TARGET text is purely factual, descriptive, or neutral, with no clear emotional stance, no judgment of people/behaviour, and no quality/value evaluation of things or events.

Examples (illustrative only):
- “The room is beautiful.” -> evaluative (1): aesthetic evaluation of a thing.
- “She was very unfair to her team.” -> evaluative (1): judgment of behaviour.
- “The concert made me really happy.” -> evaluative (1): emotional reaction.
- “The meeting starts at 3 pm.” -> non-evaluative (0): purely factual.

Important constraints:
- Do NOT invent or assume information that is not explicitly present in the TARGET text or necessary to interpret it in context. No hallucinations.
- If you are unsure, choose the more conservative option and explain briefly.

Output format:
Return ONLY a single valid JSON object (no extra text, no explanations outside JSON).

Use exactly these fields:
- "binary": 0 or 1
    - 0 = non-evaluative TARGET
    - 1 = evaluative TARGET
- "label": "no_eval" or "eval"
- "justification": a short justification for the decision (max 40 words), based ONLY on the given texts.
- "evidence_span": a short quote from the TARGET that supports the decision (or "" if none).

Context:

PREVIOUS:
\"\"\"{prev_text}\"\"\"
TARGET:
\"\"\"{target_text}\"\"\"
NEXT:
\"\"\"{next_text}\"\"\"

**Appendix 3 Prompt version 1 (Multiclass classification)**

You are an annotation assistant for Appraisal Theory (Martin & White, 2005).
The TARGET text has been identified as evaluative (binary = 1).

Task:
Classify the TARGET text into Appraisal categories. Use the three main types:
- Affect
- Judgment
- Appreciation

You are given:
- PREVIOUS: text from the row above (may be empty)
- TARGET: the current text to label
- NEXT: text from the row below (may be empty)

Use PREVIOUS and NEXT only as context (e.g., to resolve referents).
Multiple labels are allowed ONLY IF more than one label plausibly applies but none clearly dominates, mark Ambiguous = 1 and also mark the candidate labels.

Definitions (paraphrased from Appraisal Theory):

1. Affect (feelings / emotions)
  - The TARGET text expresses emotional states or reactions.
  - This can include:
    - emotional reactions to events or things (e.g., “I’m happy about it”, “That scares me”),
    - more enduring emotional dispositions (e.g., “I’m a nervous person”).
  - Typical linguistic cues:
    - emotion words (happy, sad, afraid, angry, delighted, bored, grateful, etc.),
    - phrases like “makes me feel…”, “I love / hate…”.

2. Judgment (evaluation of people and behaviour)
  - The TARGET text evaluates people or their behaviour/character in relation to social norms
    (e.g. right/wrong, fair/unfair, honest/dishonest, capable/incompetent).
  - Often about:
    - social esteem (e.g. brave, careful, talented, lazy, rude),
    - social sanction (e.g. honest, trustworthy, corrupt, unfair, immoral).
  - Typical linguistic cues:
    - adjectives for character or behaviour (“kind”, “irresponsible”, “unfair”),
    - attributions of praise or blame (“she did the right thing”, “he failed his duty”).

3. Appreciation (evaluation of things, products, events, processes, phenomena)
  - The TARGET text evaluates objects, artefacts, performances, texts, policies, situations,
    or environments in terms of quality, value, design, or aesthetics.
  - Often about:
    - reaction (how appealing or impactful something is: “boring”, “exciting”, “moving”),
    - composition (harmony, complexity, balance: “well-structured”, “chaotic”),
    - valuation (worth, significance: “important”, “valuable”, “pointless”).
  - Typical linguistic cues:
    - evaluations of things or outcomes (“beautiful view”, “poor performance”, “excellent report”).

4. Ambiguous
  - Use “ambiguous” when:
    - more than one type (Affect, Judgment, Appreciation) clearly applies,
      AND none of them is clearly dominant, OR
    - the available text is too short or vague to reliably decide which type it is.
  - In such cases, mark the types you are considering with 1 (e.g. Affect = 1, Judgment = 1)
    and set "label" to "ambiguous".

Important constraints:
- Base your decision ONLY on the given texts and the definitions above.

- Do NOT invent extra information or background that is not in the context.
- You can choose more than one type (e.g. Affect = 1 and Judgement = 1) – multi-label annotation is allowed (when you are confident that different Attitude types are expressed, as in e.g. "Despite everything, she stood tall and smiled.") This sentence explicitly shows positive Affect ("smiled") and implicitly suggests Judgement (strength, resilience, bravery).
- If you genuinely cannot decide which category is primary, use "ambiguous" and mark the candidates.
- Trust your judgment — you're interpreting how the sentence as a whole evaluates people, things, or feelings.
- Put yourself in the shoes of the speaker when trying to classify the evaluative meaning (we are interested in the speaker's opinion and assessments).

Output format:
Return ONLY a single valid JSON object (no extra text).

Use exactly these fields:
- "label": one of "affect", "judgment", "appreciation", "ambiguous"
- "affect": 0 or 1 (1 if affect is present/relevant)
- "judgment": 0 or 1 (1 if judgment is present/relevant)
- "appreciation": 0 or 1 (1 if appreciation is present/relevant)
- "probability_affect": a float between 0.0 and 1.0 for the chosen affect label in "label"
- "probability_judgment": a float between 0.0 and 1.0 for the chosen judgment label in "label"
- "probability_appreciation": a float between 0.0 and 1.0 for the chosen appreciation label in "label"
- "top_spans": a list (max 3 items) of short quotes from the TARGET that support your decision
- "explanation": a short explanation (max 60 words) of why you chose this label and flags,
explicitly linking to the definitions above.

Context:

PREVIOUS:
\"\"\"{prev_text}\"\"\"
TARGET:
\"\"\"{target_text}\"\"\"
NEXT:
\"\"\"{next_text}\"\"\"

**Appendix 4 Prompt version 2 (Multiclass classification)**
You are an annotation assistant for Appraisal Theory (Martin & White, 2005).
The TARGET text has been identified as evaluative (binary = 1).

Task:
Classify the TARGET text into Appraisal categories. Use the three main types:
- Affect
- Judgment
- Appreciation

You are given:
- PREVIOUS: text from the row above (may be empty)
- TARGET: the current text to label
- NEXT: text from the row below (may be empty)

Use PREVIOUS and NEXT only as context (e.g. to resolve referents). Multiple labels are allowed ONLY IF more than one label plausibly applies but none clearly dominates, mark Ambiguous = 1 and also mark the candidate labels.

Definitions (paraphrased from Appraisal Theory):

1. Affect (feelings / emotions)
- Put 1 if the sentence expresses feelings or emotions (e.g. happy, scared, grateful).

2. Judgment (evaluation of people and behaviour)
- Put 1 if it evaluates behaviour or character (e.g. kind, unfair, brave).

3. Appreciation (evaluation of things, products, events, processes, phenomena)
- Put 1 if it evaluates things, performance, or phenomena (e.g. beautiful, boring, excellent).

4. Ambiguous
- Put 1 if you think more than one type applies, but it is unclear which (e.g. not sure if it's Affect or Judgement). You must also mark the types you're considering (e.g. Affect = 1, Judgement = 1).

Important constraints:
- Base your decision ONLY on the given texts and the definitions above.
- Do NOT invent extra information or background that is not in the context.
- You are annotating entire sentences, not just individual words or phrases. Think about the overall meaning of the sentence (overall meaning matters, and it can be expressed explicitly or implicitly).

o Explicit: "She was terrified." → clearly Affect.

o Implicit: "She froze and couldn't speak." → may still suggest Affect (fear) even without emotion words.

- Remember that one sentence can express Appraisal/evaluation both explicitly and implicitly, e.g. "Despite everything, she stood tall and smiled." – This sentence explicitly shows positive Affect ("smiled") and implicitly suggests Judgement (strength, resilience, bravery).
- Pay attention to the co-text: sentences before and after to grasp the full meaning.
- Trust your judgment — you're interpreting how the sentence as a whole evaluates people, things, or feelings.
- Put yourself in the shoes of the speaker when trying to classify the evaluative meaning (we are interested in the speaker's opinion and assessments).

Output format:
Return ONLY a single valid JSON object (no extra text).

Use exactly these fields:
- "label": one of "affect", "judgment", "appreciation", "ambiguous"
- "affect": 0 or 1 (1 if affect is present/relevant)
- "judgment": 0 or 1 (1 if judgment is present/relevant)
- "appreciation": 0 or 1 (1 if appreciation is present/relevant)
- "probability_affect": a float between 0.0 and 1.0 for the chosen affect label in "label"
- "probability_judgment": a float between 0.0 and 1.0 for the chosen judgment label in "label"
- "probability_appreciation": a float between 0.0 and 1.0 for the chosen appreciation label in "label"
- "top_spans": a list (max 3 items) of short quotes from the TARGET that support your decision
- "explanation": a short explanation (max 60 words) of why you chose this label and flags,
explicitly linking to the definitions above.

Context:

PREVIOUS:
\"\"\"{prev_text}\"\"\"
TARGET:
\"\"\"{target_text}\"\"\"
NEXT:
\"\"\"{next_text}\"\"\"

**Appendix 5 Prompt version 3 (Multiclass classification)**

You are a research assistant.
Annotate the text according to the Appraisal Theory. Annotate each text with one of the following categories: affect, judgement, appraisal, or ambiguous.

Let's think step-by-step.

## Output Format
Return a JSON object. For a single input text, output an object with the following fields:

- "label": one of ("affect", "judgement", "appraisal", "ambiguous")
- "reasoning": a brief explanation of why this label was chosen.

For multiple input texts, output a JSON array where each entry is an object as specified above, corresponding to the input texts in order.

Example (single text):
{
  "label": "judgement",
  "reasoning": "The text expresses an evaluation of someone's behavior according to social norms."
}

Example (multiple texts):
[
  {"label": "affect", "reasoning": "The sentence communicates an emotional state."},
  {"label": "ambiguous", "reasoning": "It is unclear whether this represents an appraisal, affect, or judgement."}
]

Important constraints:
- Base your decision ONLY on the given texts and the definitions above.
- Do NOT invent extra information or background that is not in the context.
- If two or more categories are clearly present and one is clearly dominant, choose the dominant one (not ambiguous).
- If you genuinely cannot decide which category is primary, use "ambiguous" and mark the candidates.

Output format:
Return ONLY a single valid JSON object (no extra text).

Use exactly these fields:
- "label": one of "affect", "judgment", "appreciation", "ambiguous"
- "affect": 0 or 1 (1 if affect is present/relevant)
- "judgment": 0 or 1 (1 if judgment is present/relevant)
- "appreciation": 0 or 1 (1 if appreciation is present/relevant)
- "probability": a float between 0.0 and 1.0 for the chosen main label in "label"
- "top_spans": a list (max 3 items) of short quotes from the TARGET that support your decision
- "explanation": a short explanation (max 60 words) of why you chose this label and flags,
    explicitly linking to the definitions above.

Context:

PREVIOUS:
\"\"\"{prev_text}\"\"\"
TARGET:
\"\"\"{target_text}\"\"\"
NEXT:
\"\"\"{next_text}\"\"\"